\pdfoutput=1

\documentclass[11pt]{article}

\usepackage{EACL2023}

\usepackage{times}
\usepackage{latexsym}

\usepackage[T1]{fontenc}

\usepackage[utf8]{inputenc}

\usepackage{microtype}

\usepackage{inconsolata}

\usepackage{graphicx}
\usepackage{amsfonts}
\usepackage{subcaption}
\usepackage{amsmath}
\usepackage{algorithm}
\usepackage[noend]{algpseudocode}
\usepackage{algorithm}
\usepackage{multirow}
\usepackage{multicol}

\title{Task-specific Compression for Multi-task Language Models \\ using Attribution-based Pruning}

\author{Nakyeong Yang$^{1}$, Yunah Jang$^{1}$, Hwanhee Lee$^{2}$, Seohyeong Jung$^{3}$ and Kyomin Jung$^{1}$ \\
  $^{1}$Seoul National University,
  $^{2}$Chung-Ang University,
  $^{3}$Hyundai Motor Group and 42dot Inc\\
  \texttt{\{yny0506, vn2209, kjung\}@snu.ac.kr}\\
  \texttt{hwanheelee@cau.ac.kr}, \texttt{seohyeong.jeong@42dot.ai}
  }

\begin{document}

\maketitle
\begin{abstract}
Multi-task language models show outstanding performance for various natural language understanding tasks with only a single model.
However, these language models utilize an unnecessarily large number of model parameters, even when used only for a specific task.
This paper proposes a novel training-free compression method for multi-task language models using a pruning method.
Specifically, we use an attribution method to determine which neurons are essential for performing a specific task. 
We task-specifically prune unimportant neurons and leave only task-specific parameters.
Furthermore, we extend our method to be applicable in low-resource and unsupervised settings.
Since our compression method is training-free, it uses few computing resources and does not destroy the pre-trained knowledge of language models.
Experimental results on the six widely-used datasets show that our proposed pruning method significantly outperforms baseline pruning methods.
In addition, we demonstrate that our method preserves performance even in an unseen domain setting.
\end{abstract}

\section{Introduction}\label{sec:intro}
Various pre-trained language models with large-scale data and parameters have emerged \citep{devlin2018bert,lewis2019bart,raffel2019exploring,brown2020language}.
Specifically, pre-trained language models like T5 \citep{raffel2019exploring} and GPT-3 \citep{brown2020language} have shown outstanding performance on many natural language understanding tasks.
These language models can perform various tasks with a single model by treating every text processing problem as a text generation problem.
However, these language models may utilize unnecessary large-scale model parameters even when performing only a specific task.
Previous works have introduced various compression methods for language models such as pruning \citep{chen2020earlybert, goyal2020power, he2021magic}, knowledge distillation \citep{sanh2019distilbert, hou2020dynabert, DBLP:journals/corr/abs-2004-04124, sun2020mobilebert}, quantization \citep{shen2020q}, and low-rank factorization \citep{DBLP:journals/corr/abs-2103-11367}.
However, these studies have (1) not compressed the language models task-specifically or (2) demanded an additional training process like the case of knowledge distillation.
This additional training process requires excessive computing resources and a massive training dataset.
Furthermore, this training process can destroy inherent pre-trained knowledge in language models since it updates the model's pre-trained parameters \cite{toneva2018empirical}.
Due to the catastrophic forgetting \cite{mccloskey1989catastrophic} caused by pre-trained knowledge destruction, models which are compressed and trained for a specific task, tend to show degraded performance on solving other pre-trained tasks \cite{kirkpatrick2017overcoming, ritter2018online}.
Also, additional memory space is required to store the trained parameters separately.

\begin{figure*}[!ht]
    \centering
    \centerline{\includegraphics[width=440pt]{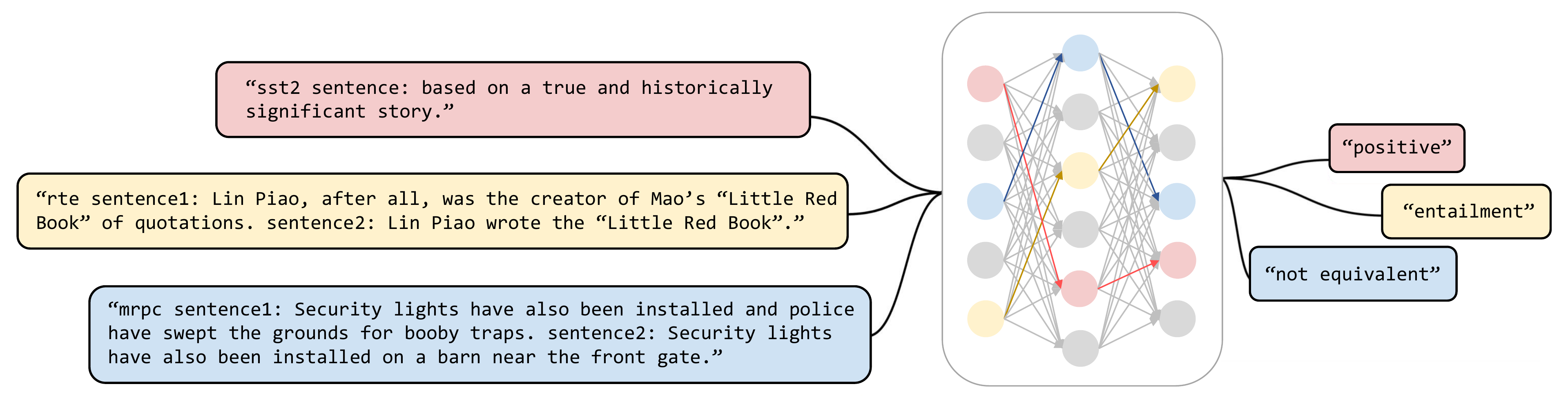}}
    \caption{Task-specific Knowledge of Multi-task Language Models. Not all parameters in a language model behave as important parameters when performing a single task. For example, in this figure, when a language model receives SST-2 data, sentiment analysis data, only the parameters expressed in red color behave as essential parameters.}
    \vspace{-0.4cm}
    \label{fig:intro}
\end{figure*}

In this paper, we propose a novel training-free attribution-based task-specific pruning method that enables more efficient compression and inference by extracting only task-specific parameters from multi-task language models.
We can determine which neurons are essential to derive a specific output for each neural network layer by using attribution so that we can extract only task-specific parameters from the entire model, as shown in Figure~\ref{fig:intro}.
We can efficiently process input data while preserving the model's task performance by selecting only the important neurons determined by the attribution method.
Furthermore, we extend our method to be applicable in two challenging scenarios: low-resource and unsupervised scenarios.
The former alleviates insufficient labeled data situations, and the latter handles settings when labels are unavailable.
Both methods can relieve the cost of obtaining labeled datasets, which requires excessive human resources and is time-consuming.
Especially under the low-resource setting, our attribution-based task-specific pruning requires only a single forward and backward propagation computation for few-shot data samples (e.g., only ten samples) to derive attribution of each neuron.
Since this pruning process does not update the model's parameters, it does not destroy the pre-trained knowledge of the language models.
Therefore, it is irrelevant to the various disadvantages that arise during an additional training process.
Since our method is model-agnostic, it can be applied to any neural network model broadly and generally.
Even we can use it to extract only task-specific knowledge after other compression methods are applied.

Experimental results on the six widely-used natural language understanding tasks show that our proposed method significantly outperforms baseline training-free pruning methods.
Furthermore, we demonstrate that our method shows robust performance in both low-resource and unsupervised settings.
Also, we reveal that our proposed method shows outstanding knowledge preservation even for an unseen related domain, which suggests that our method can preserve task-specific knowledge effectively.
We additionally investigate to offer a guideline for our task-specific compression method by analyzing which types of layers are significant for processing task-specific knowledge.

\section{Related Works}
\subsection{Efficient Language Models}
As transformer-based \cite{vaswani2017attention} 
language models \cite{devlin2018bert, radford2018improving, raffel2019exploring, liu2019roberta, yang2019xlnet} have become state-of-the-arts on many NLP tasks in the last few years, deep neural network model compression methods have been vastly applied to large-scale language models. \citet{fan2019reducing} randomly drops layers at training time, which enables structured pruning on transformer layers at inference time. \citet{michel2019sixteen} prunes less important attention heads at inference time. Other works \cite{goyal2020power, kim2021learned} focus on pruning less important tokens and progressively remove them during inference. However, many of the pruning methods \cite{goyal2020power, kim2021learned, chen2020earlybert} require a following fine-tuning step of the model parameters after fixing the configuration of a pruned network, which makes such methods undesirable for efficient task-specific compression.

On the knowledge distillation side, \citet{sun2019patient, jiao2019tinybert, sanh2019distilbert} employ teacher-student framework \cite{hinton2015distilling} to transfer knowledge from an original large model (teacher), to a lightweight shallow model (student). They differ in how the student network is initialized and to which components knowledge distillation is applied. On the other hand, \citet{shen2020q} uses the mixed precision group-wise quantization based on Hessian information to compress BERT. 

There are other streams of works that explore efficient language models by solving the bottleneck of the Transformer-based model computation. \citet{beltagy2020longformer} and \citet{zaheer2020big} sparsify the attention matrix to make transformer-based language models more efficient and \citet{wang2020linformer} applies low-rank approximation to increase inference speed. However, such works sparsify the full self-attention matrix according to attention score, which does not directly reduce the dimension of the matrices in the model such as query, key, value, and feed-forward matrices.

\subsection{Network Pruning}
One of the ways to categorize network pruning is to compare structured pruning to unstructured pruning. For structured pruning ~\cite{li2016pruning, hu2016network, wen2016learning}, groups of weight connections are removed from a network together, such as entire channels or filters in CNN-based networks and layers or attention heads in transformer-based networks. For unstructured pruning~\cite{han2015deep, han2015learning}, weight connections are removed from a network individually. However, unstructured pruning methods produce large sparse weight matrices which are computationally inefficient unless equipped with a specifically designed hardware. In this paper, we utilize the structured pruning method to propose a compression method that enables efficient weight matrix multiplication computation.

\subsection{Attribution Method}
We utilize an attribution method \citep{shrikumar2016not} to extract the importance of neurons from the pre-trained language models.
Attribution methods are mostly used to derive important features \textit{(i.g., pixel, token)} to extract interpretability from deep neural networks \citep{baehrens2010explain,springenberg2014striving,shrikumar2016not}.
Specifically, attribution methods are used to compute the importance of each feature for performing a specific task.
Formally, suppose we have a function $\mathcal{P}:\mathbb{R}^{d}\rightarrow[0,1]^{m}$ that represents deep neural networks for multi-class classification.
The contribution of the $i$-th feature in $x$ to the prediction of $c$-th class using $\mathcal{P}$ is defined as follows:

\begin{equation}
\begin{aligned}
    A^{(x,c)}_{i}(x)=x_{i}\times \frac{\partial \mathcal{P}(c|x)}{\partial x_{i}}
\end{aligned}
\label{eq:attr}
\end{equation}

\noindent where $\partial \mathcal{P}(c|x)/\partial x_{i}$ is the gradient of $\mathcal{P}(c|x)$ with respect to the $i$-th feature.

\section{Methodologies}
In this section, we describe our attribution-based pruning method for extracting only the task-specific knowledge from a multi-task language model T5 \cite{raffel2019exploring}, where attribution is obtained using gradient information. Furthermore, we extend our method to low-resource and unsupervised settings to alleviate insufficient labeled data situations. We select T5 because it is a multi-task solving model and can be used in any natural language understanding setting by treating every text processing problem as a text generation problem. For our problem setting, suppose we have input text $x=\{x_{1}, ..., x_{n}\}$ and output text $y=\{y_{1}, ..., y_{m}\}$ mapped as $(x, y) \in\mathcal{D}$, where each text corresponds to a sequence of tokens, and an input text contains a prefix task description.
We can represent a standard conditional language modeling objective to maximize the following likelihood:

\begin{equation}
\begin{aligned}
    \mathcal{L}(x, y)=\sum_{i}\log \mathcal{P}(y_{i}|x,y_{1},...,y_{i-1};\Theta)
\end{aligned}
\label{eq:lm}
\end{equation}

\noindent where the conditional probability $\mathcal{P}$ is modeled using a neural network with parameters $\Theta$.

\begin{figure*}[!ht]
\newcommand\x{3.5}
\newcommand\widthPercent{0.5}
\begin{subfigure}[t]{\widthPercent\textwidth}
    \centering
    \includegraphics[width=\x cm]{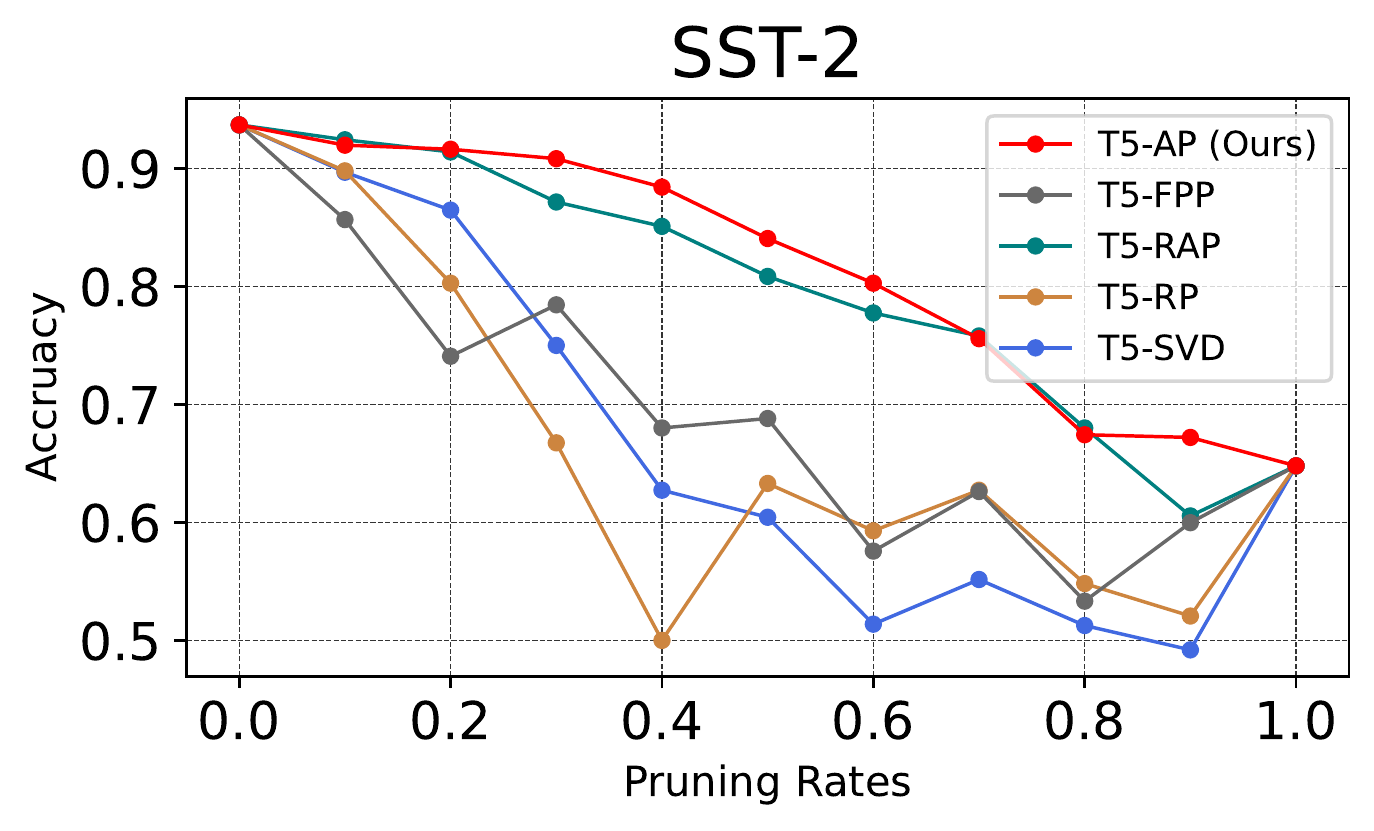}
    \includegraphics[width=\x cm]{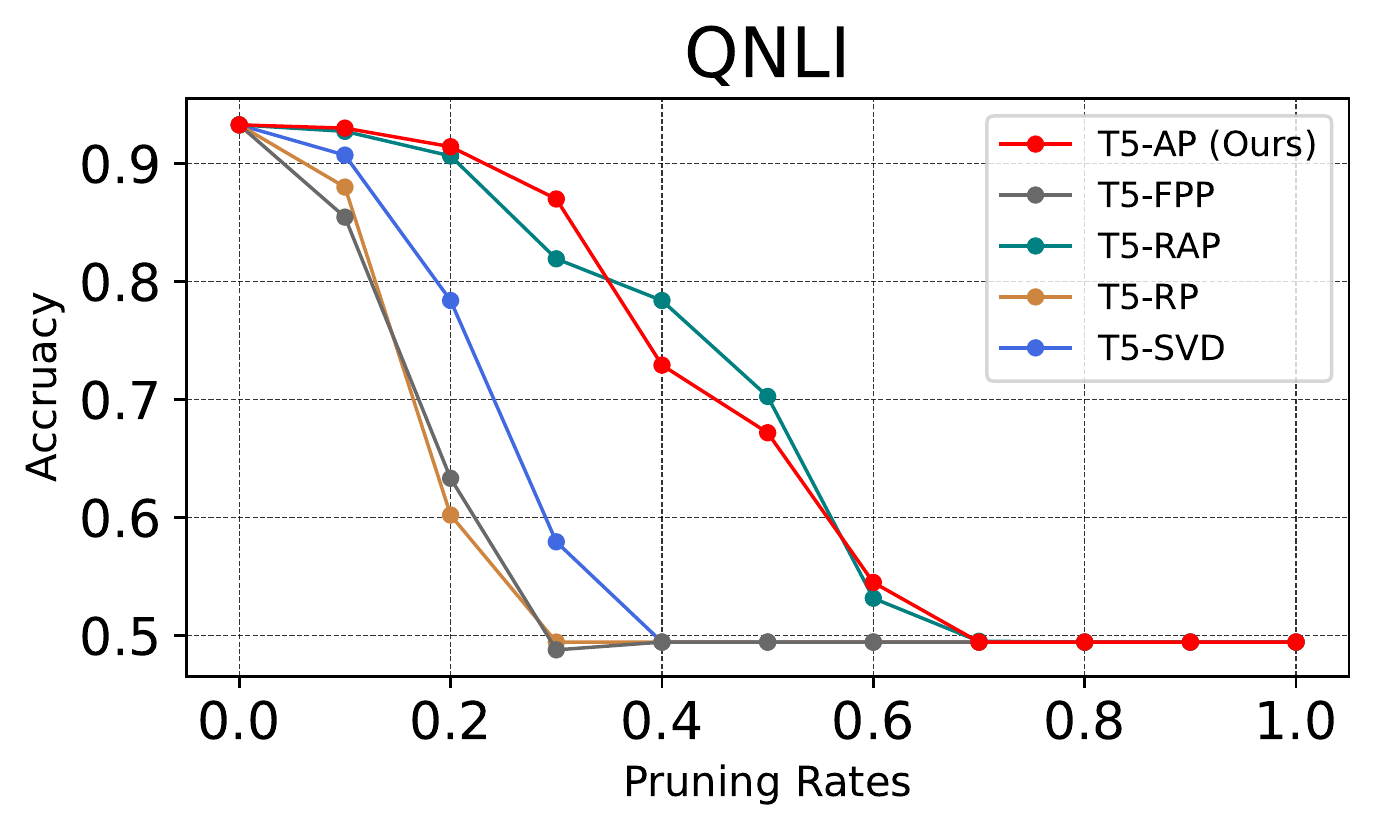}
\end{subfigure}
\begin{subfigure}[t]{\widthPercent\textwidth}
    \centering
    \includegraphics[width=\x cm]{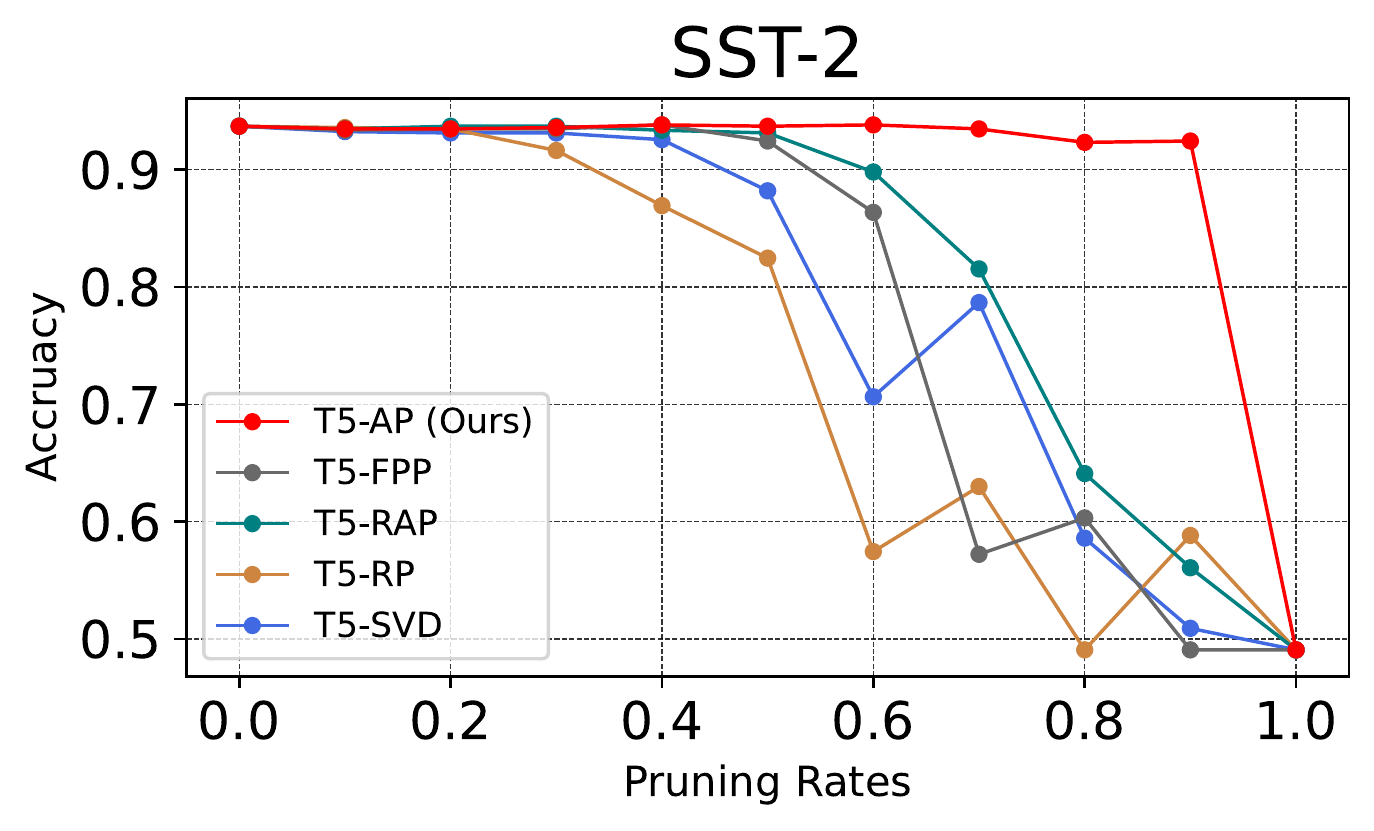}
    \includegraphics[width=\x cm]{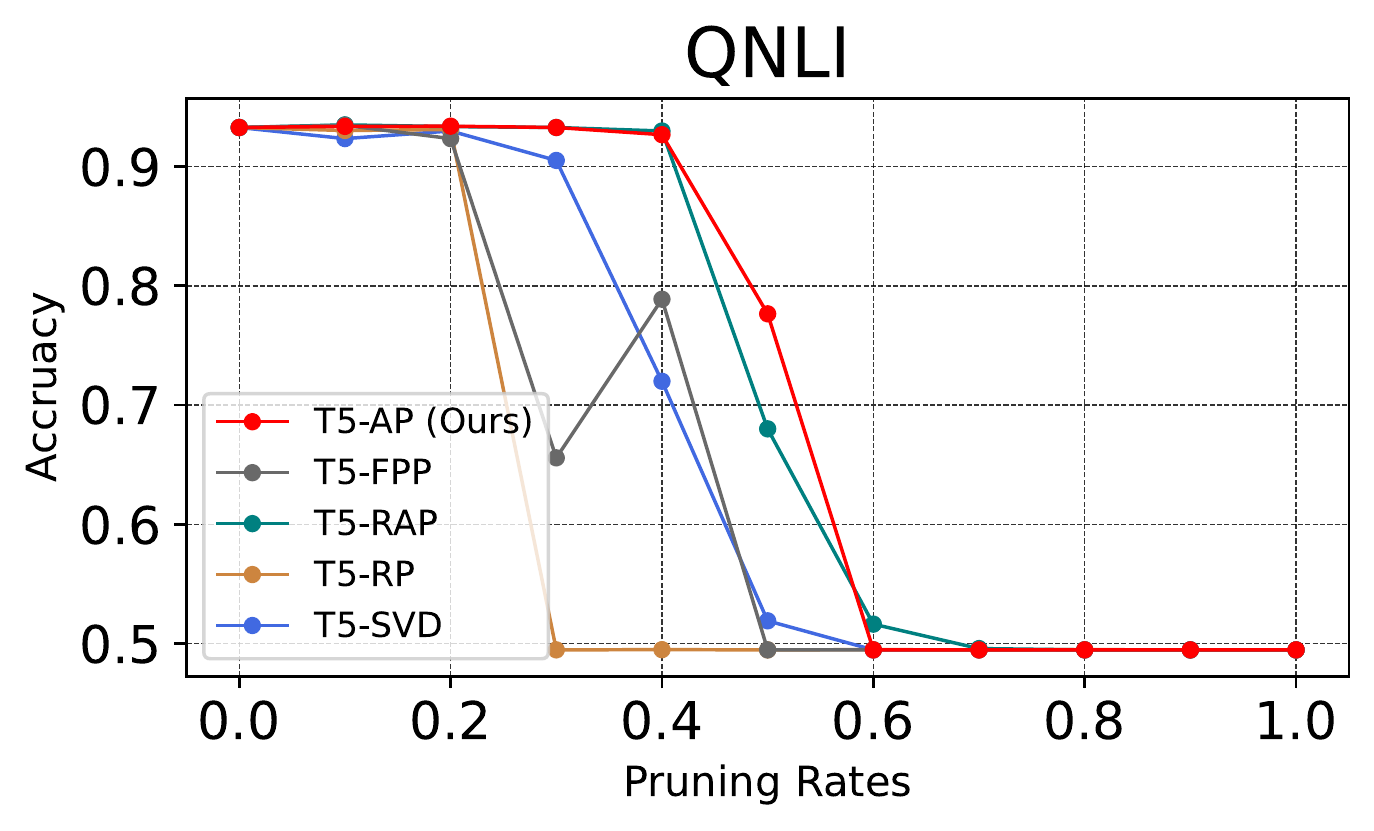}
\end{subfigure}

\begin{subfigure}[t]{\widthPercent\textwidth}
    \centering
    \includegraphics[width=\x cm]{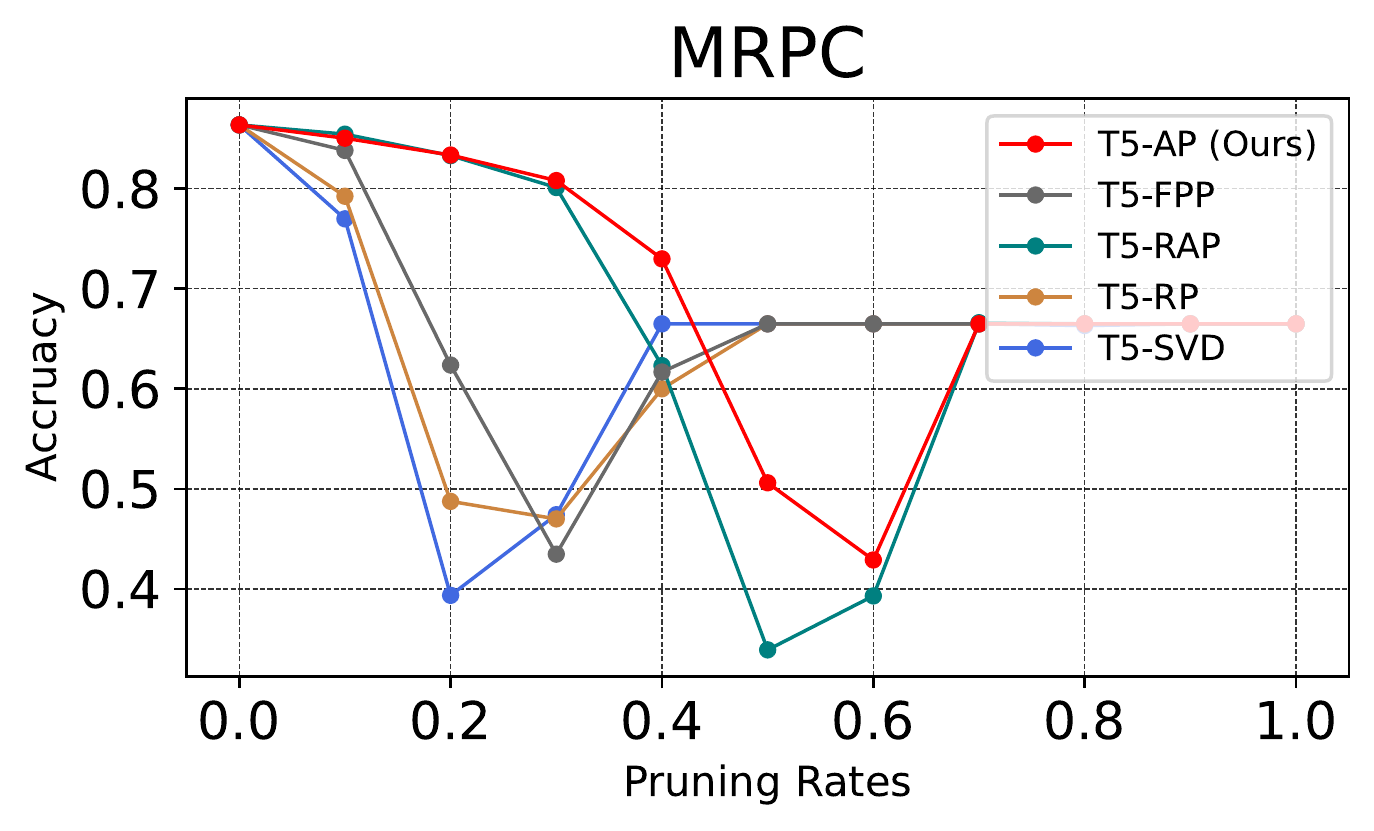}
    \includegraphics[width=\x cm]{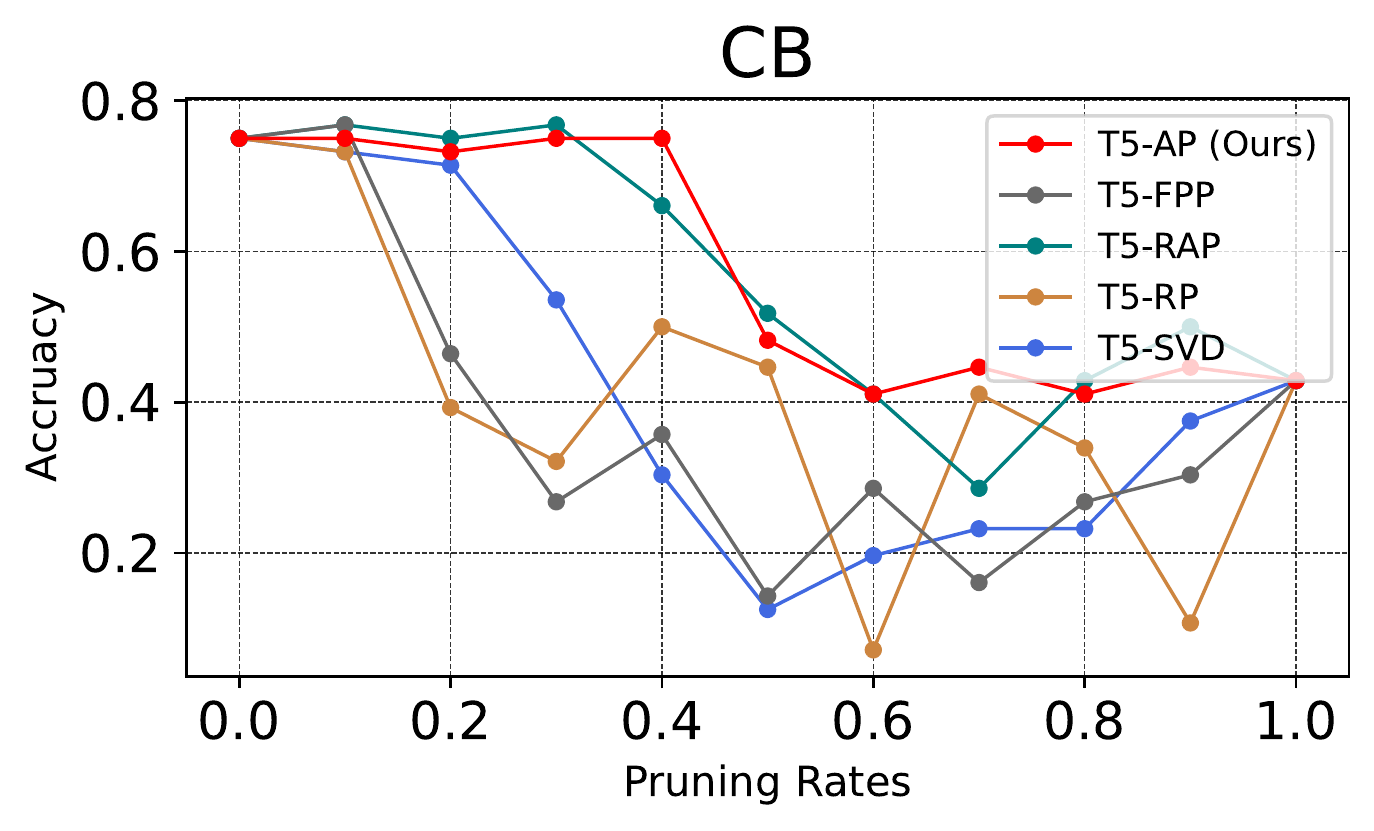}
\end{subfigure}
\begin{subfigure}[t]{\widthPercent\textwidth}
    \centering
    \includegraphics[width=\x cm]{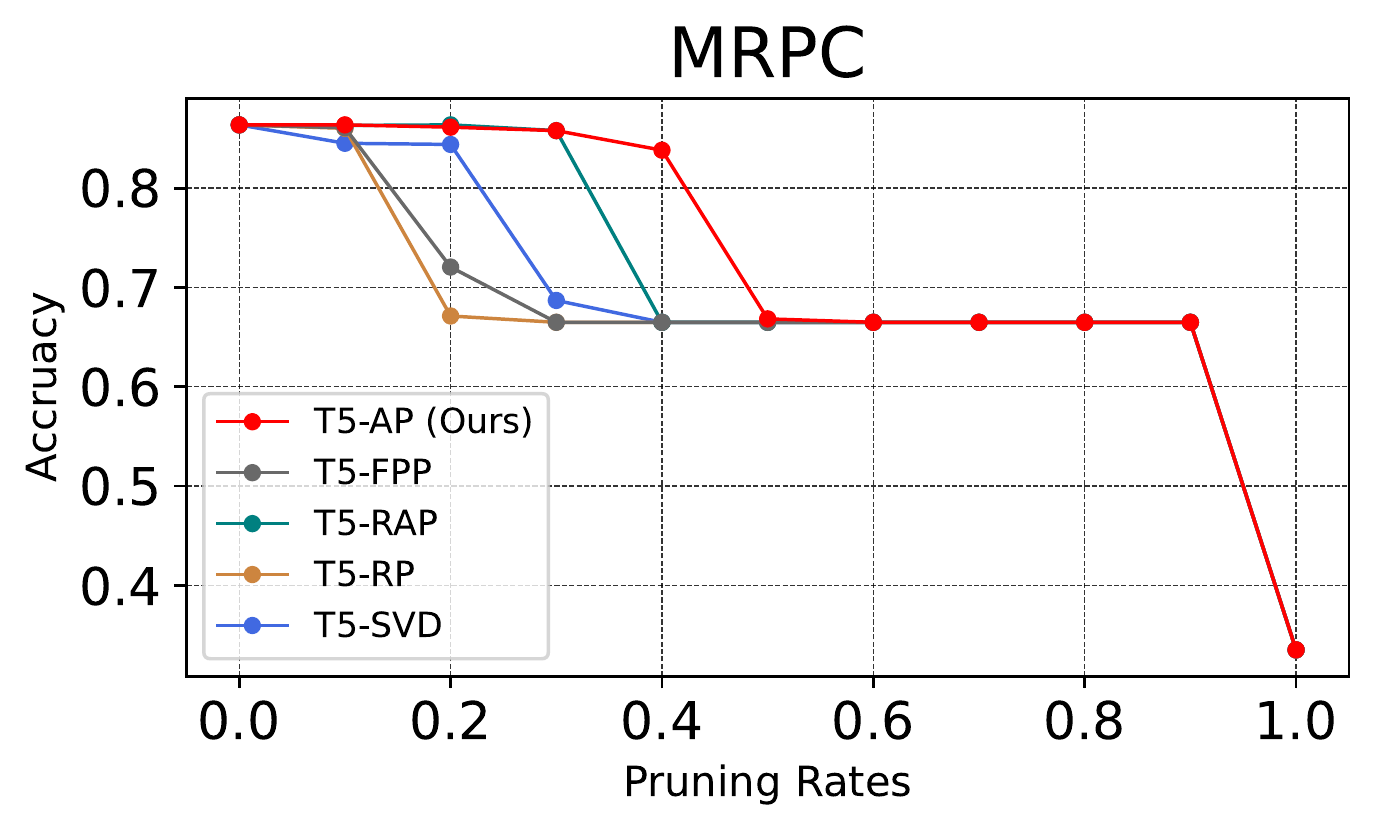}
    \includegraphics[width=\x cm]{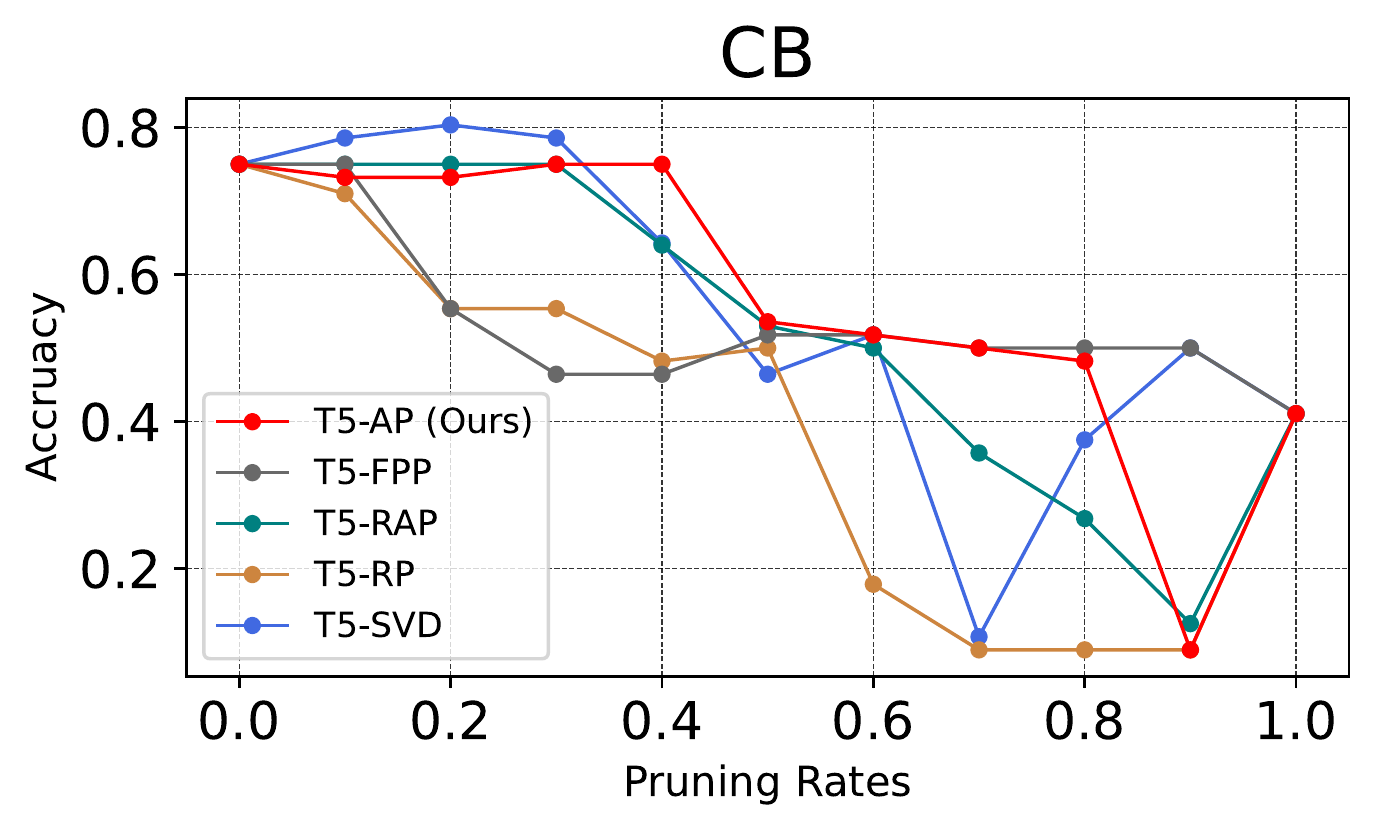}
\end{subfigure}

\begin{subfigure}[t]{\widthPercent\textwidth}
    \centering
    \includegraphics[width=\x cm]{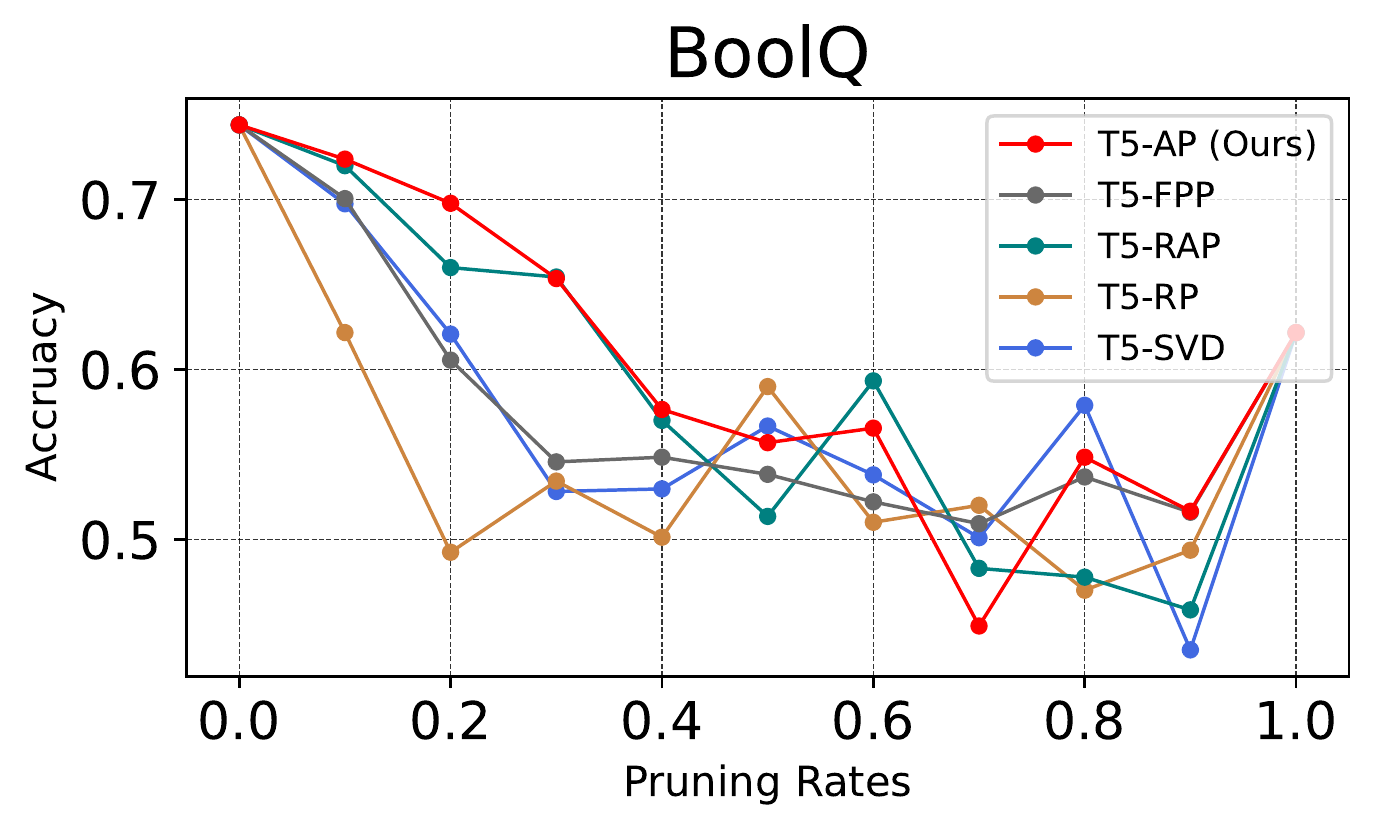}
    \includegraphics[width=\x cm]{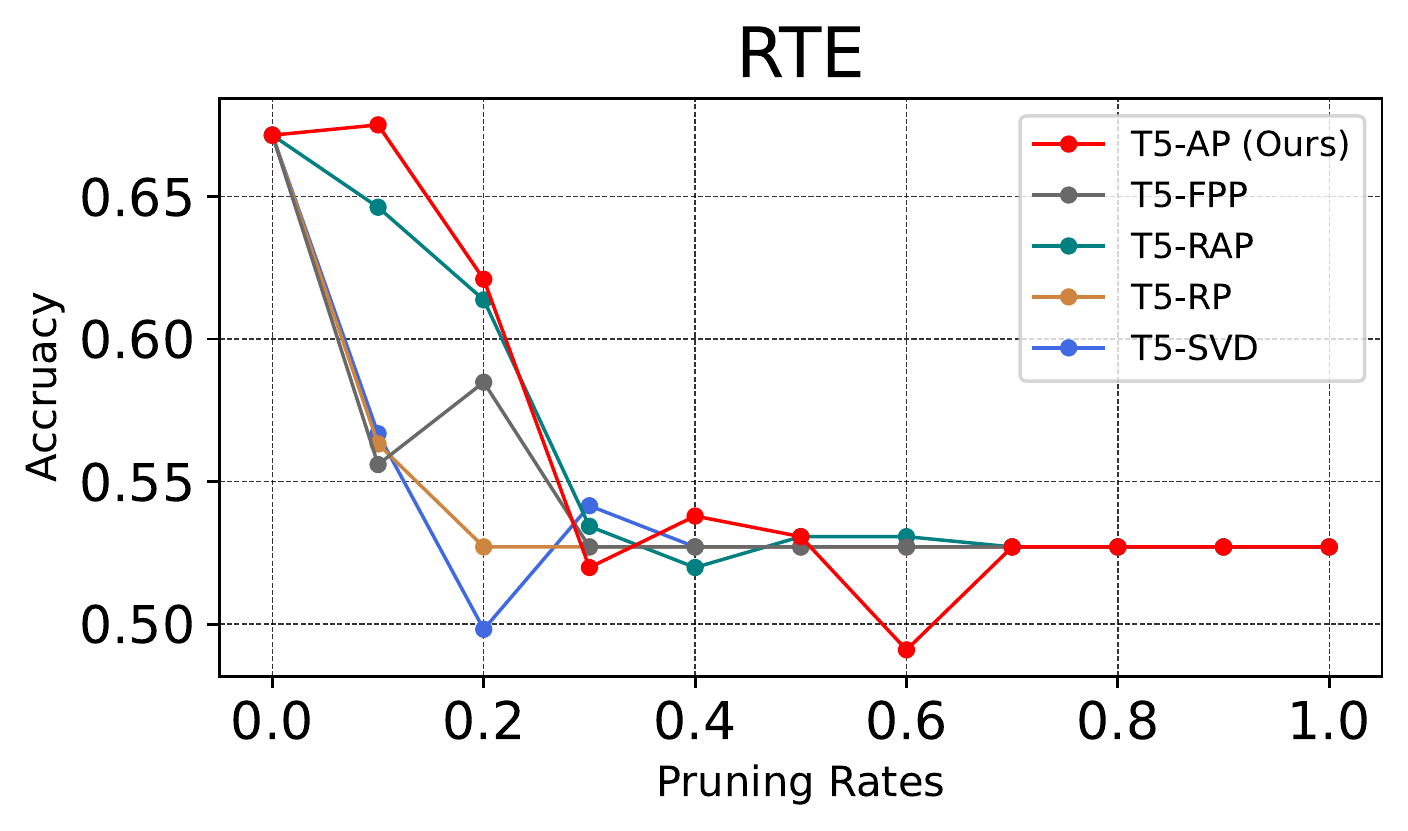}
    \caption{Encoder Pruning Results}
\end{subfigure}
\begin{subfigure}[t]{\widthPercent\textwidth}
    \centering
    \includegraphics[width=\x cm]{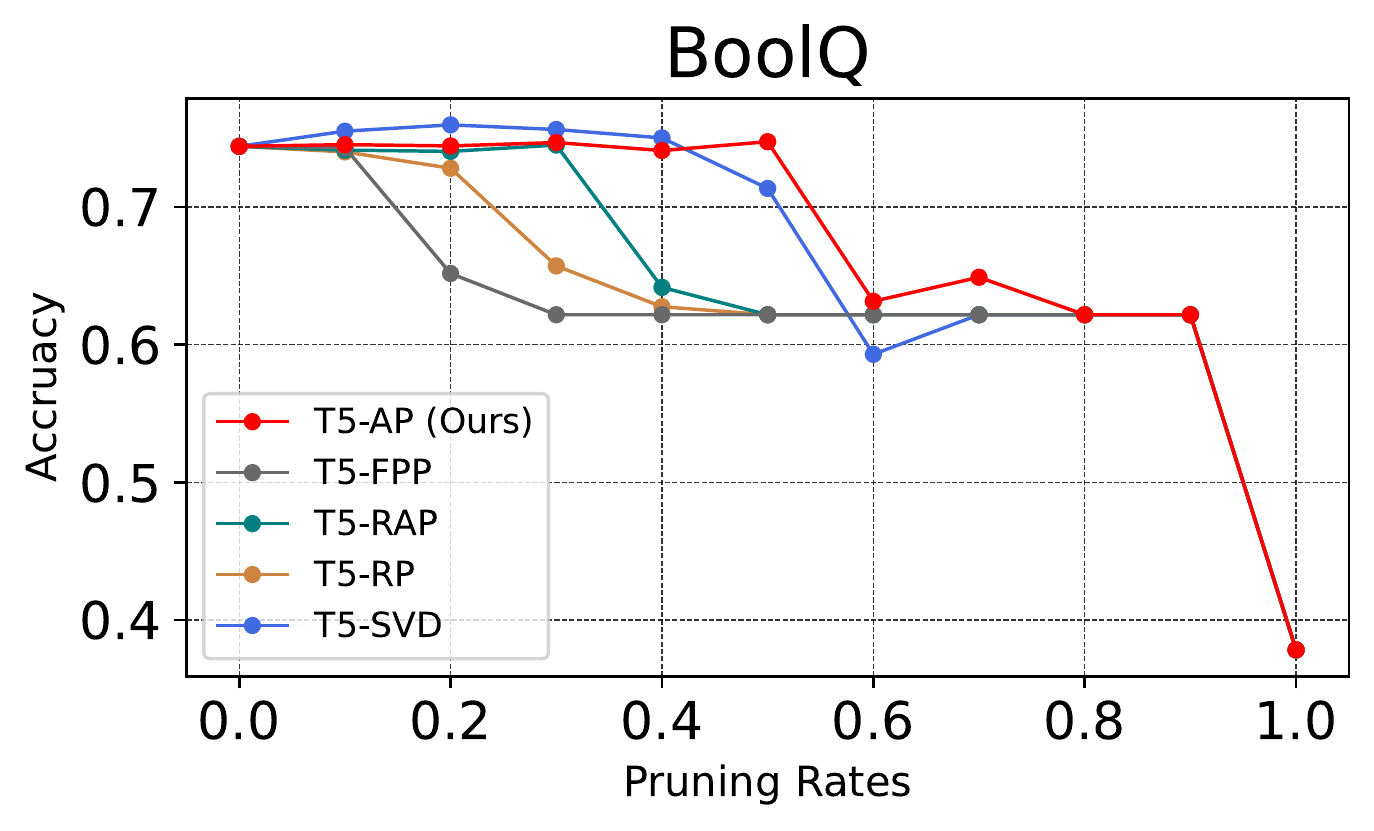}
    \includegraphics[width=\x cm]{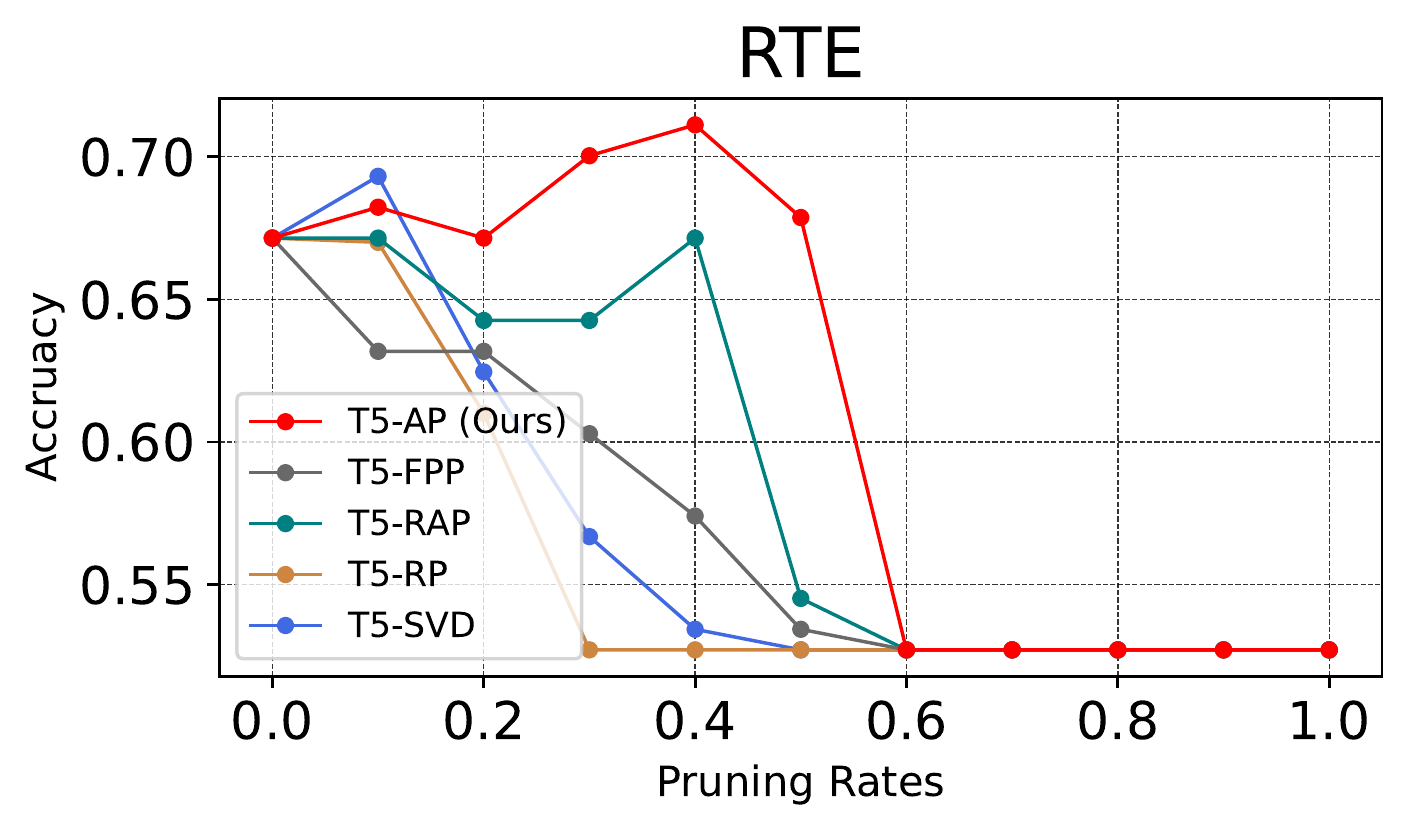}
    \caption{Decoder Pruning Results}
\end{subfigure}

\caption{Module-specific Pruning Results. Our proposed attribution-based pruning significantly outperforms the other pruning methods in most cases. Especially, our task-specific pruning is more effective on decoder compression; these results suggest that most task-specific knowledge exists in the decoder of language models. The standard deviations of T5-RP and T5-RAP are shown in appendix~\ref{sec:appendix2}.}
\label{fig:independent}
\vspace{-0.3cm}
\end{figure*}

\subsection{Task-specific Knowledge Extraction}
\paragraph{Applying Pruning for Transformer variants}
Deep neural networks can be compressed by pruning unimportant $i$-th neurons of the layer representation $h$ \citep{han2015deep,han2015learning}.
The architecture of Transformer-based models mainly consists of multi-head attentions and fully connected feed-forward networks as follows.

\footnotesize
\begin{equation}
    \begin{aligned}
        MultiHead(Q,K,V) = Concat(head_{1}, ..., head_{h})W^{O}\\
        head_{i} = Attention(QW_{i}^{Q}, KW_{i}^{K}, VW_{i}^{V})\:\:\:\:\:\:\:\:\:\:\\\
        FFN(x) = \sigma(xW_{1}+b_{1})W_{2} + b_{2} \:\:\:\:\:\:\:\:\:\:\:\:\:\:\:\:\:\
    \end{aligned}
    \label{eq:attn_ffn}
\end{equation}
\normalsize

where $W_{i}^{Q,K,V}\in{\mathbb{R}^{d_{model}\times{d_{q,k,v}}}}$ and $W_{i}^{O}\in{\mathbb{R}^{d_{v}\times{d_{model}}}}$ are the projection matrix parameters for multi-head attentions. For the fully connected feed-forward network (FFN), two linear transformations, denoted with the projection matrix parameters $W_{1}$ and $W_{2}$ and biases $b_{1}$ and $b_{2}$, with an activation function are used.
Transformer \citep{vaswani2017attention} variants can be compressed by pruning $W^{Q,K,V,O}$, $W_{1,2}$, and $b_{1,2}$ for each transformer block.

\paragraph{Deriving Attribution for Language Models}
Language models generate text outputs by iteratively selecting a word-piece from the vocabulary dictionary.
Therefore, the text generation process can be seen as a classification task dealt with in the attribution methods, and we can apply the attribution methods to compute the importance of features for language models.
However, the purpose of this study is to derive the importance of each neuron $h_i$ in the layer representation $h \in \mathbb{R}^{d}$, rather than deriving the importance for the input feature $x_i$.
Hence, the attribution formula is adapted to compute a neuron attribution $A^{(x,y_{j})}_{i}\in \mathbb{R}$ as follows:

\begin{equation}
\begin{aligned}
    A^{(x,y_{j})}_{i}(h)=h_{i}\times \frac{\partial \mathcal{P}(y_{j}|x, y_{1:j-1})}{\partial h_{i}} \\
\end{aligned}
\label{eq:attr_lm1}
\end{equation}

If the target output text consists of multiple word-pieces rather than a single word-piece, language models must derive the multiple word-piece output distributions.
Therefore, we change the attribution formula to handle multiple word-piece outputs as follows:

\begin{equation}
\begin{aligned}
    A^{(x,y)}_{i}(h)=h_{i}\times \sum_{j=1}^{|y|} \frac{\partial \mathcal{P}(y_{j}|x, y_{1:j-1})}{\partial h_{i}}
\end{aligned}
\label{eq:attr_lm2}
\end{equation}

Since $A^{(x,y)}_{i}$ is attribution for one sample data $x$, we obtain the final neuron attribution by summing attributions for multiple sample data as shown in the following formula:

\begin{equation}
\begin{aligned}
    A^{(\mathcal{D})}_{i}(h)=\sum_{(x,y) \in \mathcal{D}} A^{(x,y)}_{i}(h)
\end{aligned}
\label{eq:attr_lm3}
\end{equation}

\noindent where $\mathcal{D}$ means the entire task-specific dataset.
In low-resource environments, few-shot samples can be used for $\mathcal{D}$ (e.g., only ten samples), which are sufficient to derive a precise importance score for each neuron.
Experimental results for low-resource setting are described in section~\ref{sec:experiments-low-resource}.

\paragraph{Attribution-based Layer Pruning}
We focus on applying attribution-based pruning on the Transformer encoder and decoder, more specifically on multi-head attention and fully connected feed-forward networks. 
We use neuron attribution $A^{(\mathcal{D})}_i$ as the importance for each neuron of a specific layer.
We sort the importance of each neuron in order of magnitude at each layer, and we can compress the model by pruning neurons with lower importance.

\begin{equation}
\small
\begin{aligned}
    argsort_{i}(A)=&|\{j|(A_{i}<A_{j}) \cup (A_{i}=A_{j}, j<i)\}| \\ 
    & \textbf{where}\ i,j \in \{1,...,k\}
\end{aligned}
\end{equation}
\label{eq:pruning1}
\noindent Once neurons are sorted according to the importance score, we prune neurons from each layer with the pruning rate $p$ by constructing a set $\mathcal{M}$ of neuron indices to be secured.

\begin{equation}
\small
\begin{aligned}
    \mathcal{M} = \{i | argsort_{i}(A) < \lfloor k\times p \rfloor\} \\
    \textbf{where}\ i \in \{1,...,k\} \:\:\:\:\:\:\:\:\:\:\:\
\end{aligned}
\label{eq:pruning2}
\end{equation}

\noindent The algorithm for deriving a set $\mathcal{M}$ is shown in appendix~\ref{sec:appendix1}.
Suppose $W \in \mathbb{R}^{d \times k}$ is a linear matrix multiplication parameter we want to prune, the matrix after pruning is denoted as $\tilde{W} = (W_{ij})_{\substack{1\leq i\leq d\\ j\in \mathcal{M}}}$.
If the bias term $b \in \mathbb{R}^{k}$ is added to the operation for an affine transformation, the bias term can also be compressed by performing the $\tilde{b} = (b_{i})_{i \in \mathcal{M}}$ operation similarly.
The compressed parameters are used to compute the new representation by performing the transformation operation $h\tilde{W}$ or $h\tilde{W}+\tilde{b}$.

More specifically, for $W_i^{Q}$, $W_i^{K}$, and $W_i^{V}$ from eq. \eqref{eq:attn_ffn}, second dimension (the number of columns) of the matrix is pruned and for $W_i^{O}$, $W_{1}$, and $W_{2}$, the first dimension (the number of rows) is pruned to preserve the original architecture by matching shape with input processed from the previous layer. After pruning, multi-head attention and fully connected feed-forward network computations are precisely the same as before but with the pruned weight matrices:

\footnotesize
\begin{equation}
    \begin{aligned}
        MultiHead(Q,K,V) = Concat(head_{1}, ..., head_{h})\tilde{W}^{O}\\
        head_{i} = Attention(Q\tilde{W}_{i}^{Q}, K\tilde{W}_{i}^{K}, V\tilde{W}_{i}^{V})\:\:\:\:\:\:\:\:\:\:\\\
        FFN(x) = \sigma(x\tilde{W}_{1}+b_{1})\tilde{W}_{2} + \tilde{b}_{2}\:\:\:\:\:\:\:\:\:\:\:\:\:\:\:\:\:\
    \end{aligned}
    \label{eq:attn_ffn_pruned}
\end{equation}
\normalsize

\noindent Note that attribution scores are sorted locally within each layer, and the pruning rate $p$ is applied to each prunable layer uniformly.

Our proposed compression process utilizes a structured pruning without any training process.
Therefore, our method can conduct on-demand real-time task-specific compression and inference for each task while preserving pre-trained parameters.
The detailed algorithm for on-demand real-time task-specific compression and inference is shown in appendix~\ref{sec:appendix1}.

\begin{figure*}[!ht]
\newcommand\x{3.4}
\newcommand\y{1.3}
\begin{minipage}[b]{\textwidth}
    \centering
    \centerline{
    \includegraphics[width=\x cm]{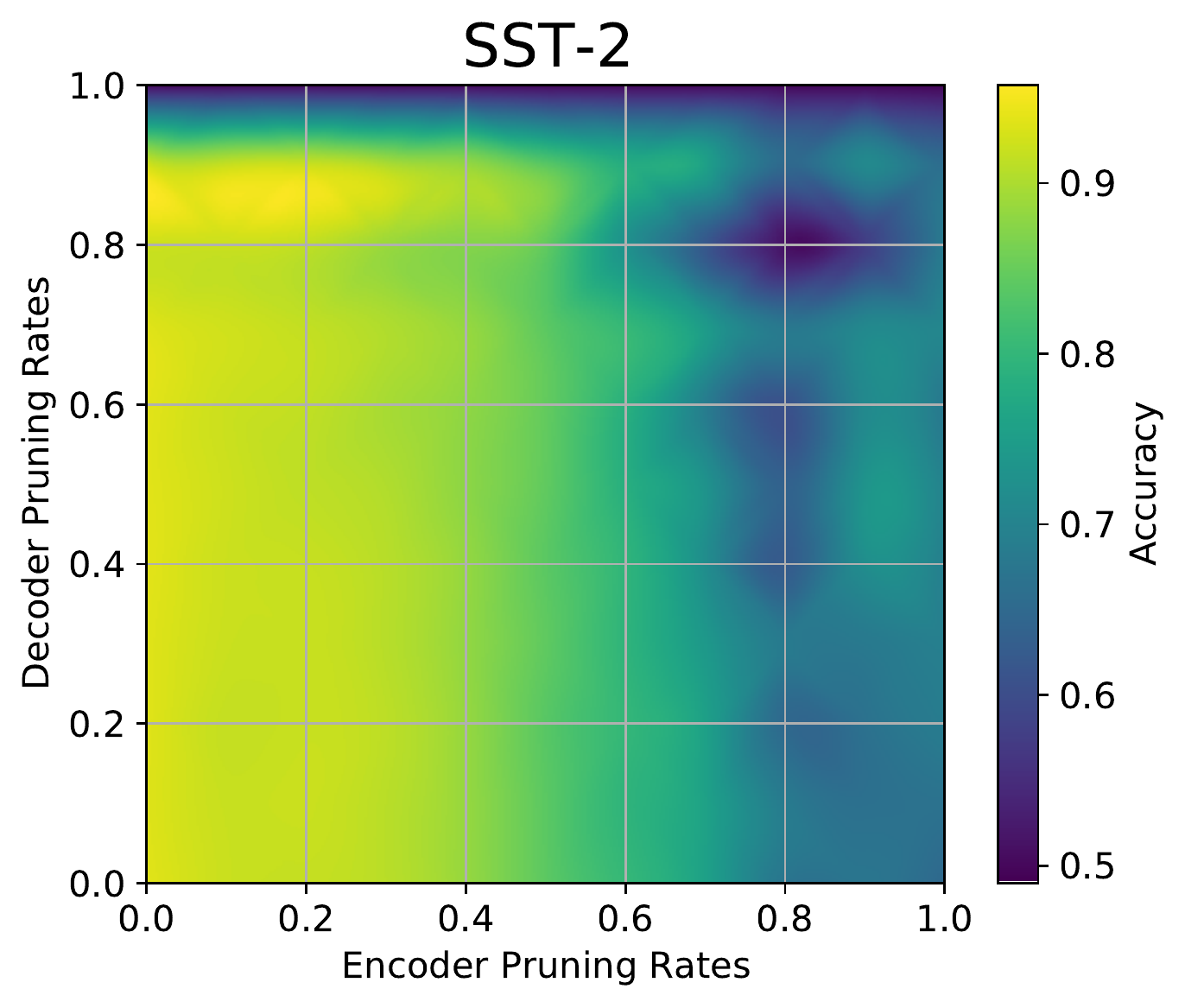}
    \hspace{\y cm}
    \includegraphics[width=\x cm]{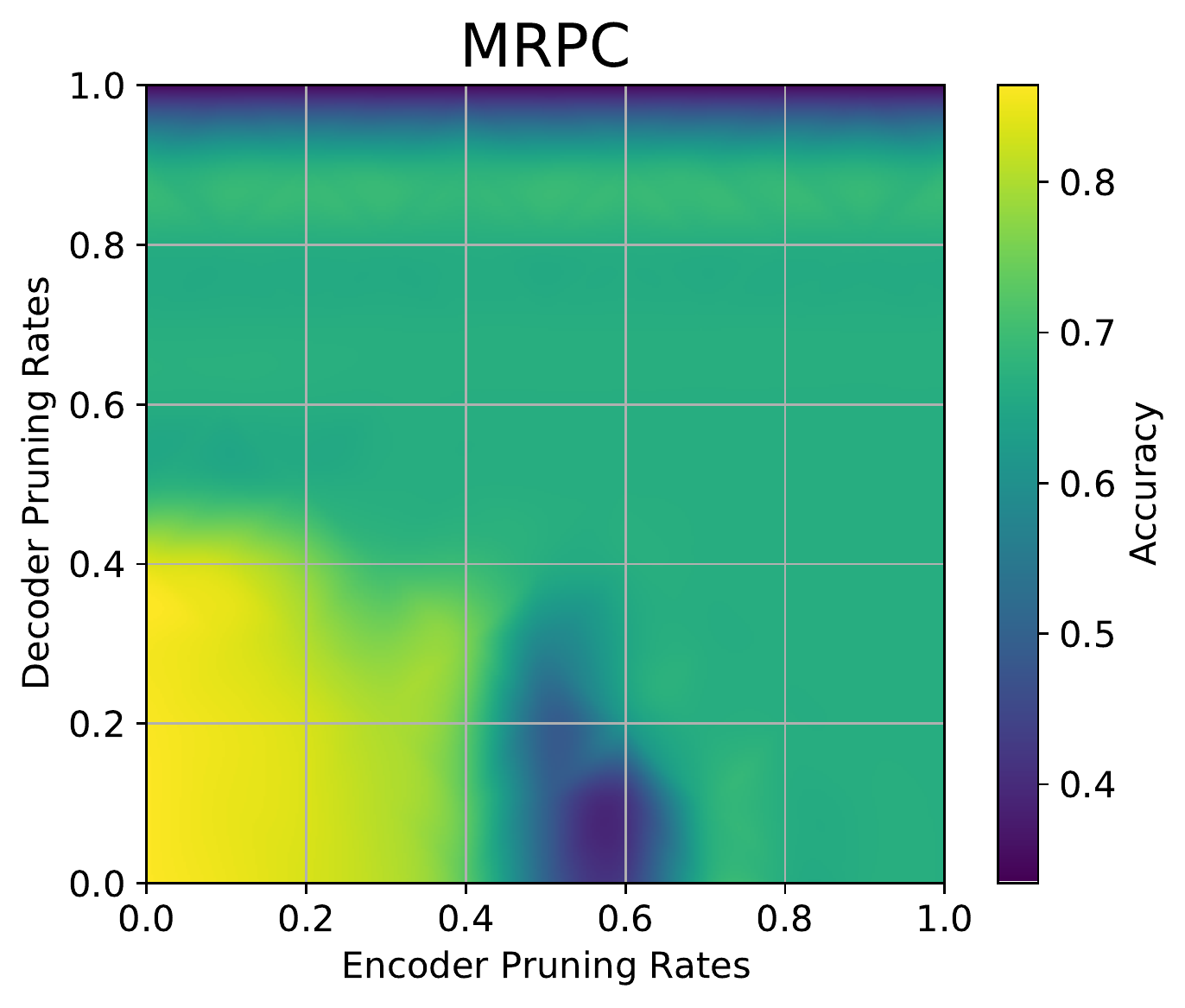}
    \hspace{\y cm}
    \includegraphics[width=\x cm]{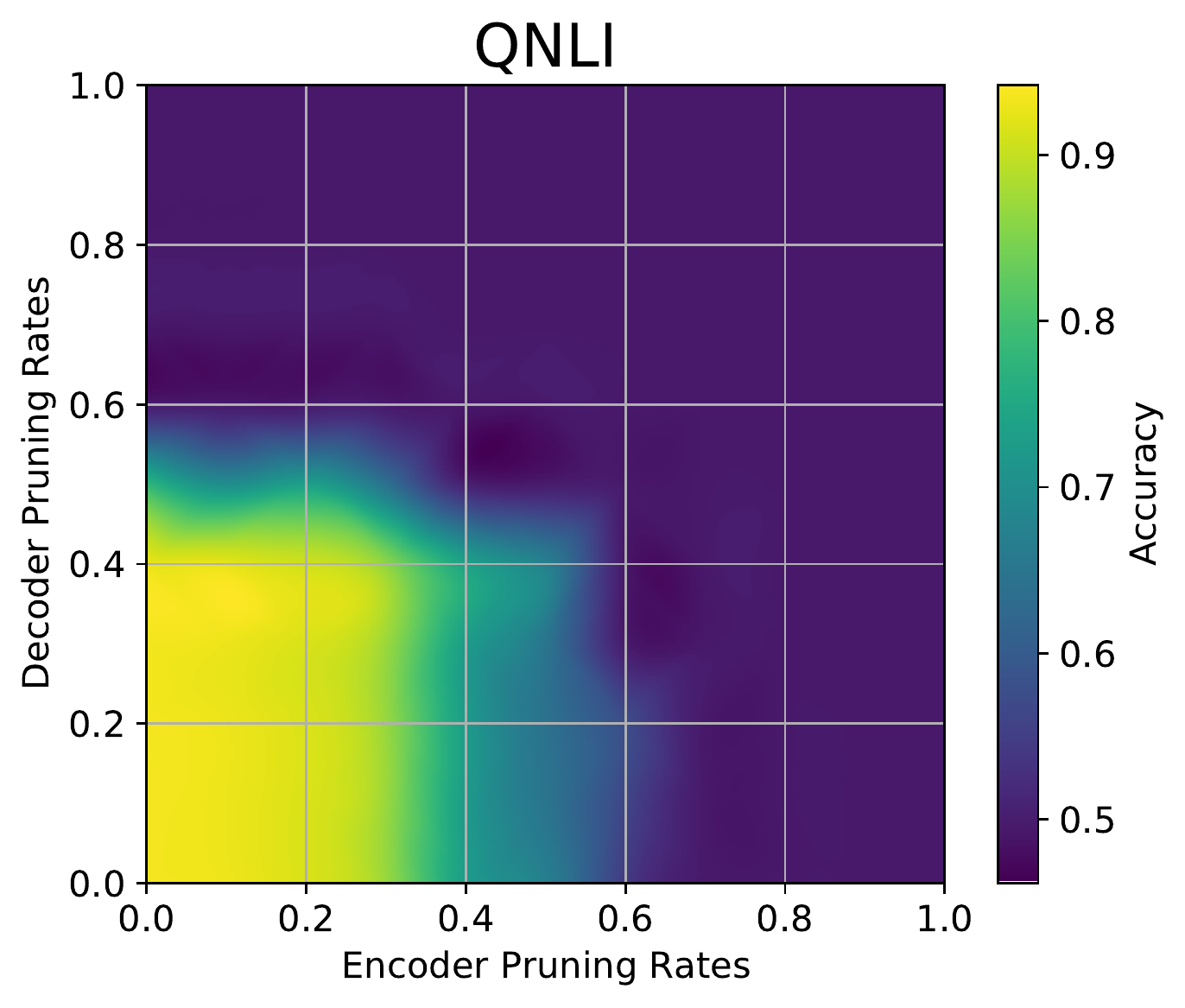}
    }
\end{minipage}

\begin{minipage}[b]{\textwidth}
    \centering
    \centerline{
    \includegraphics[width=\x cm]{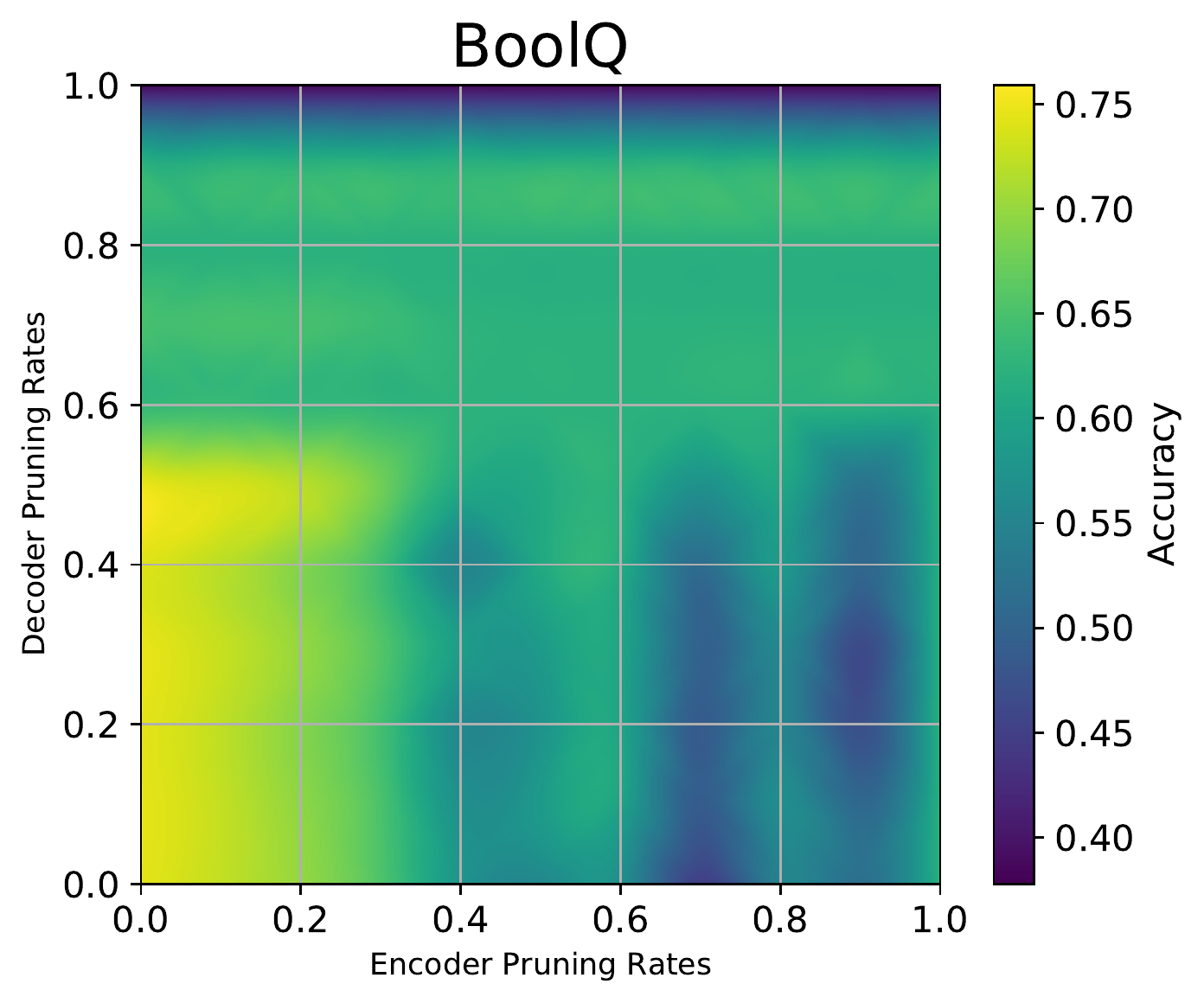}
    \hspace{\y cm}
    \includegraphics[width=\x cm]{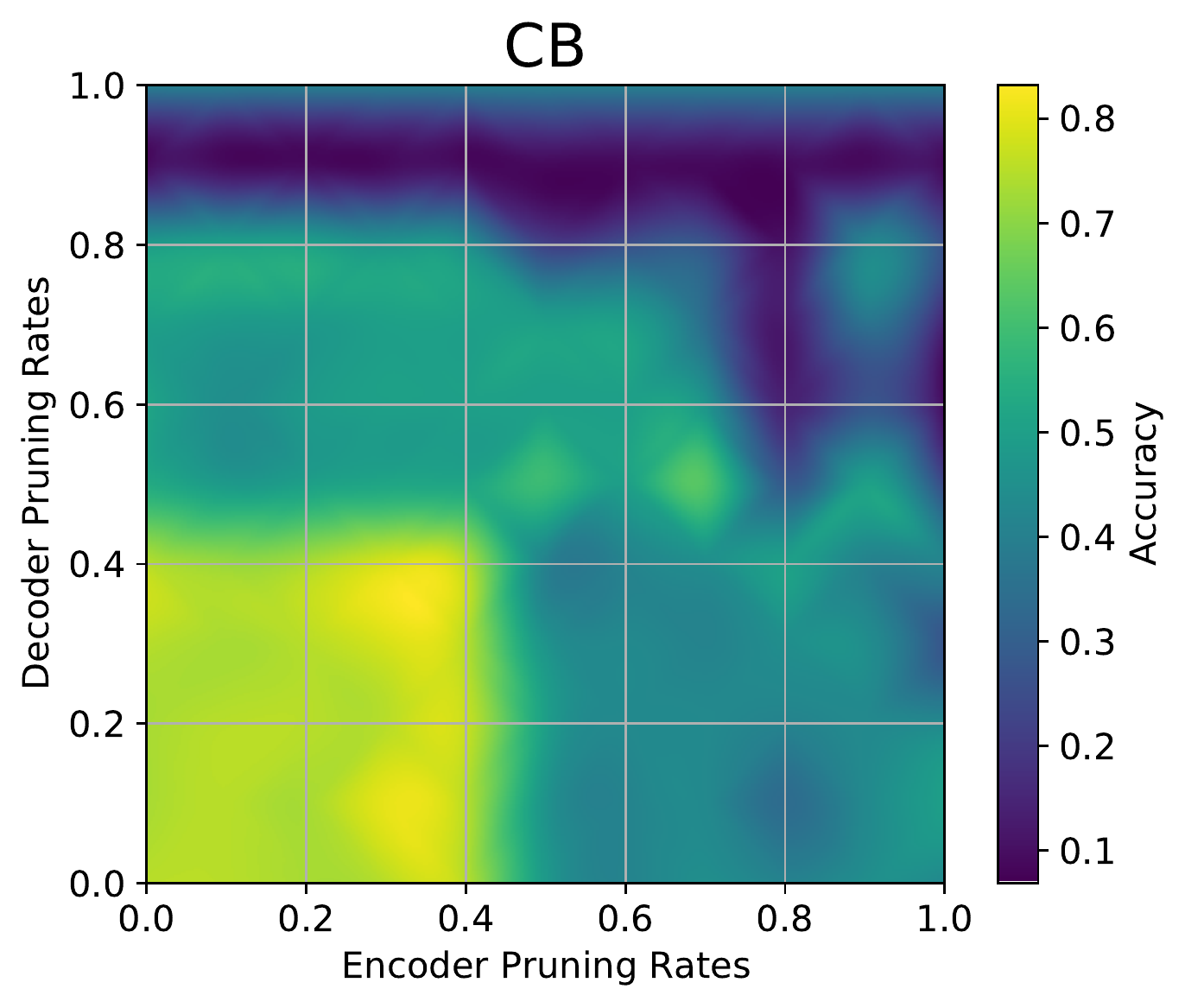}
    \hspace{\y cm}
    \includegraphics[width=\x cm]{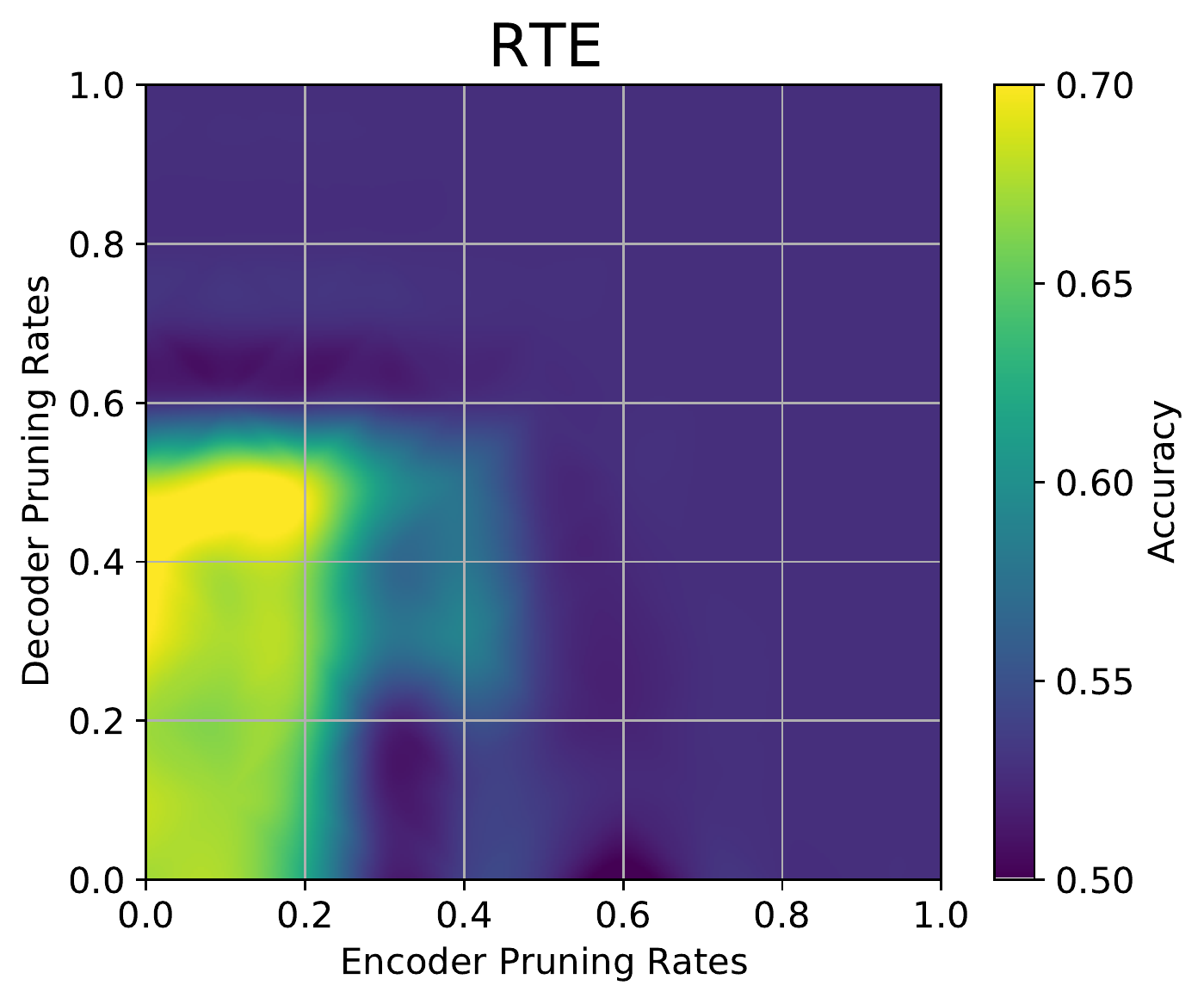}
    }
\end{minipage}
\vspace{-0.5cm}
\caption{Module-integrated Pruning Results. These results reveal that compressing the whole architecture of the model does not additionally degrade the model’s performance compared to module-specific pruning. We experiment with the combinations of ten pruning rates for the encoder and decoder, and plot the interpolated results.}
\label{fig:integrated}
\vspace{-0.45cm}
\end{figure*}

\subsection{Unsupervised Pruning}
\label{sec:method-unsupervised}
Obtaining labeled data usually requires excessive human resources and is time-consuming.
Therefore, we propose an additional method to derive attributions in an unsupervised setting to mitigate this problem.
If the label for the dataset is given, we can simply compute attribution by summing the gradients values for the word-piece set composing the label.
However, when the label is not given, the target word-piece set is ambiguous.
To resolve this problem, we compute task-specific importance by summing the absolute values of attributions for all candidate labels as follows:

\begin{equation}
\begin{aligned}
    A^{(x,\mathcal{Y})}_{i}(h)=\sum_{y \in \mathcal{Y}} |h_{i}\times \sum_{j=1}^{|y|} \frac{\partial \mathcal{P}(y_{j}|x, y_{1:j-1})}{\partial h_{i}}|
\end{aligned}
\label{eq:unsupervised}
\end{equation}

\noindent where $\mathcal{Y}$ is the candidate label set.
The above importance computation formula does not require supervision for any data.
Hence, we may not reflect definite label information when computing each neuron's importance under our unsupervised compression setting.
However, this setting is helpful for a resource-constrained environment, where obtaining labeled data is challenging.

\section{Experiments}
\subsection{Experimental Setup}
\paragraph{Datasets}
We conduct experiments on six downstream tasks \citep{wang2018glue, wang2019superglue}.
Specifically, we utilize SST-2 (sentiment analysis); MRPC (semantic textual similarity); BoolQ (question answering); and QNLI, CB, RTE (natural language inference).
\paragraph{Implementation Details}
We select pre-trained \textit{T5-base}\footnote{https://huggingface.co/t5-base} as a backbone for the following experiments.
\textit{T5-base} consists of 12 encoder and 12 decoder layers.
Each encoder layer contains 6 prunable matrices: 4 for the multi-head self-attention networks and 2 for the feed-forward networks. Each decoder layer contains 10 prunable matrices: 4 for the multi-head self-attention networks and 2 for the feed-forward networks, and 4 for the cross-attention networks.
\textit{T5-base} used in our experiments has been fine-tuned by multi-task learning using the six datasets above.
We experiment with pruning rates ranging from 0.1 to 1.0, and a pruning rate is applied to each prunable layer uniformly.

\subsection{Task-specific Pruning Efficiency}
In this section, we validate the effectiveness of our task-specific attribution-based pruning by comparing the performance with other pruning methods.
We collect compressed models using various pruning methods and evaluate the model's performance on testset for all six datasets.

\paragraph{Baselines}
We select four other training-free pruning methods to compare with our task-specific \textbf{T5 Attribution Pruning (T5-AP)}.
\begin{itemize}
  \item \textbf{T5 Forward Propagation Pruning (T5-FPP)} derives the importance of each neuron with the absolute value of the forward propagation value of each neuron.
  This method is widely used to compress model in various studies \citep{han2015learning,hu2016network,li2016pruning}.
  Previous studies using FPP generally fine-tune the compressed model to increase the model's performance.
  However, we eliminate the fine-tuning process to maintain a fair evaluation scenario since we focus on studying training-free compression.
  \item \textbf{T5 Low Rank Factorization (T5-SVD)} prunes weight matrices of neural networks using Singular Value Decomposition (SVD).
  SVD is commonly used as a main matrix compression idea in various researches  \citep{wang2020linformer, noach2020compressing}.
  Specifically, SVD is used to compress a matrix based on low rank factorization formula as follows:
  
  \begin{equation}
    \begin{aligned}
        W = U\Sigma V \approx \sum_{j=1}^{r} \sigma_{j} \times (U_{j} \times V_{j})
    \end{aligned}
    \label{eq:svd1}
  \end{equation}
  
  where $W \in \mathbb{R}^{d\times k}$ is a matrix to compress, and $U \in \mathbb{R}^{d\times r}$ and $V \in \mathbb{R}^{r\times k}$ are the decomposed matrices.
  $\Sigma=diag(\sigma_{1}, \sigma_{2}, ..., \sigma_{r})$ is a diagonal matrix consisting of the singular values $\sigma_{i}$, where $r\leq min(d, k)$ is the matrix rank. $U_{j}$ is the $j$-th column of $U$ and $V_{j}$ is the $j$-th row of $V$.
  We can compress the matrices of T5 by determining the rank $r = \lfloor \frac{d \times k \times p}{d+k+1} \rfloor$ to have the same number of parameters as T5-AP, where $p$ is the pruning rate defined in formula~\ref{eq:pruning2}.
  
  
  \item \textbf{T5 Random Attribution Pruning (T5-RAP)} randomly selects word-pieces that are not label, and uses them to compute attribution.
  RAP does not derive appropriate task-specific importance for each neuron since this method randomly selects word-pieces output.
  We calculate the final performance of T5-RAP by averaging the accuracy derived from five trials of random word-pieces selection.
  \item \textbf{T5 Random Pruning (T5-RP)} randomly selects which neuron to prune.
  This method can achieve the lower-bound performance of overall training-free pruning methods since it randomly selects which neuron to prune without any knowledge.
  We calculate the final performance of T5-RP by averaging the accuracy derived from five trials of random pruning.
\end{itemize}

\paragraph{Module-specific Pruning}
For each dataset, we separately compressed the encoder and decoder at varying pruning rates to reveal the effect of our method on the encoder and decoder, respectively.
Figure~\ref{fig:independent} shows the experimental results for five compression methods, including our proposed method.
Experimental results show that our method outperforms other compression methods in most cases.
Specifically, there is almost no performance difference between the T5-RP and T5-FPP.
These results suggest that the T5-FPP does not extract task-specific knowledge.
In addition, T5-SVD performs not badly in some cases, but generally performs similarly to T5-RP.
It is because the low-rank approximation of T5-SVD does not work task-specifically.
Surprisingly, T5-RAP sometimes performs similarly to T5-AP, probably due to the use of partial gradients information calculated from model parameters.
Our experiments show that the decoder part of T5 has the robustness for task-specific compression than the encoder part of T5.
These results demonstrate that T5 decoder processes more task-specific information than T5 encoder.

\begin{figure*}[!ht]
\newcommand\x{3.3}
\begin{subfigure}[t]{0.5\textwidth}
    \centering
    \includegraphics[width=\x cm]{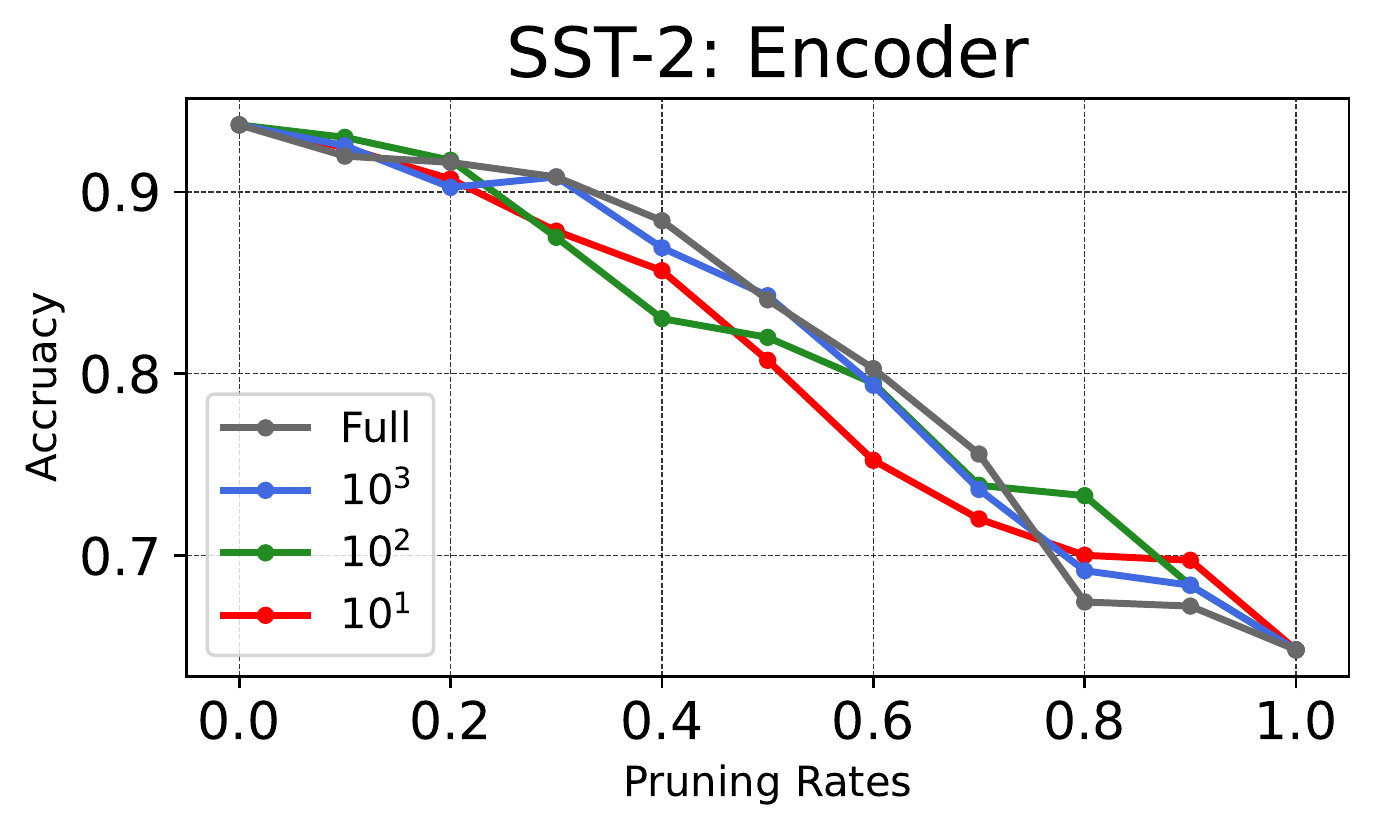}
    \includegraphics[width=\x cm]{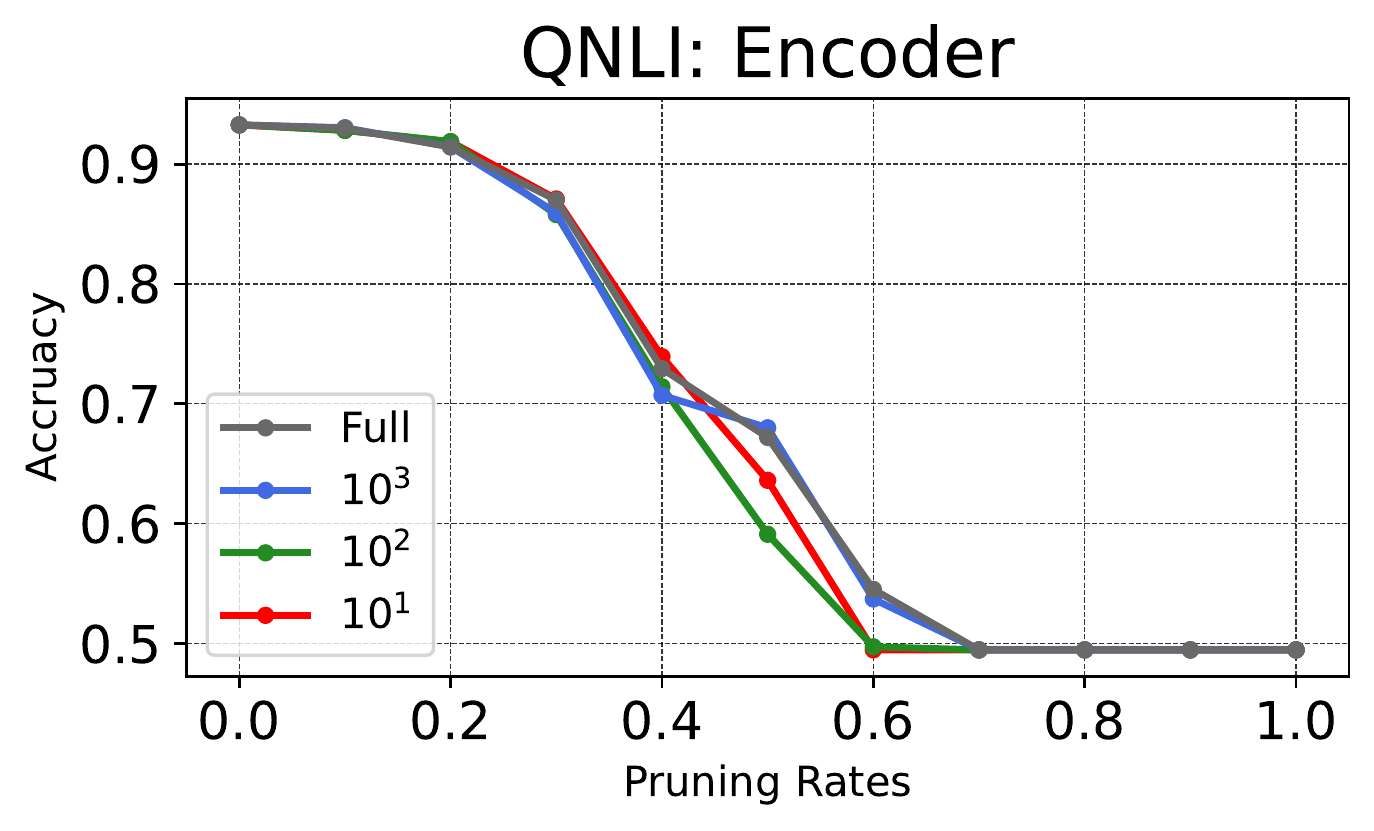}
\end{subfigure}
\begin{subfigure}[t]{0.245\textwidth}
    \centering
    \includegraphics[width=\x cm]{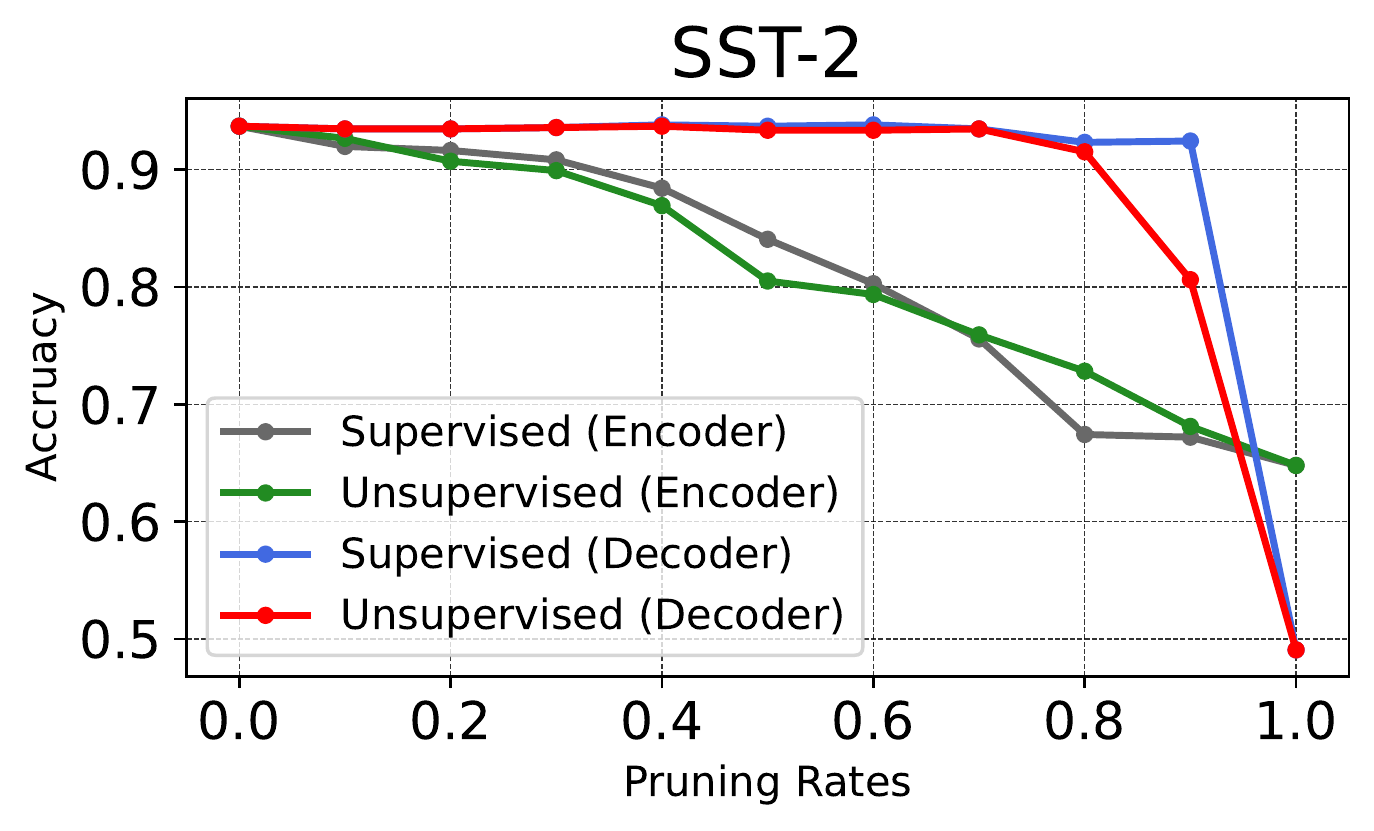}
\end{subfigure}
\begin{subfigure}[t]{0.245\textwidth}
    \centering
    \includegraphics[width=\x cm]{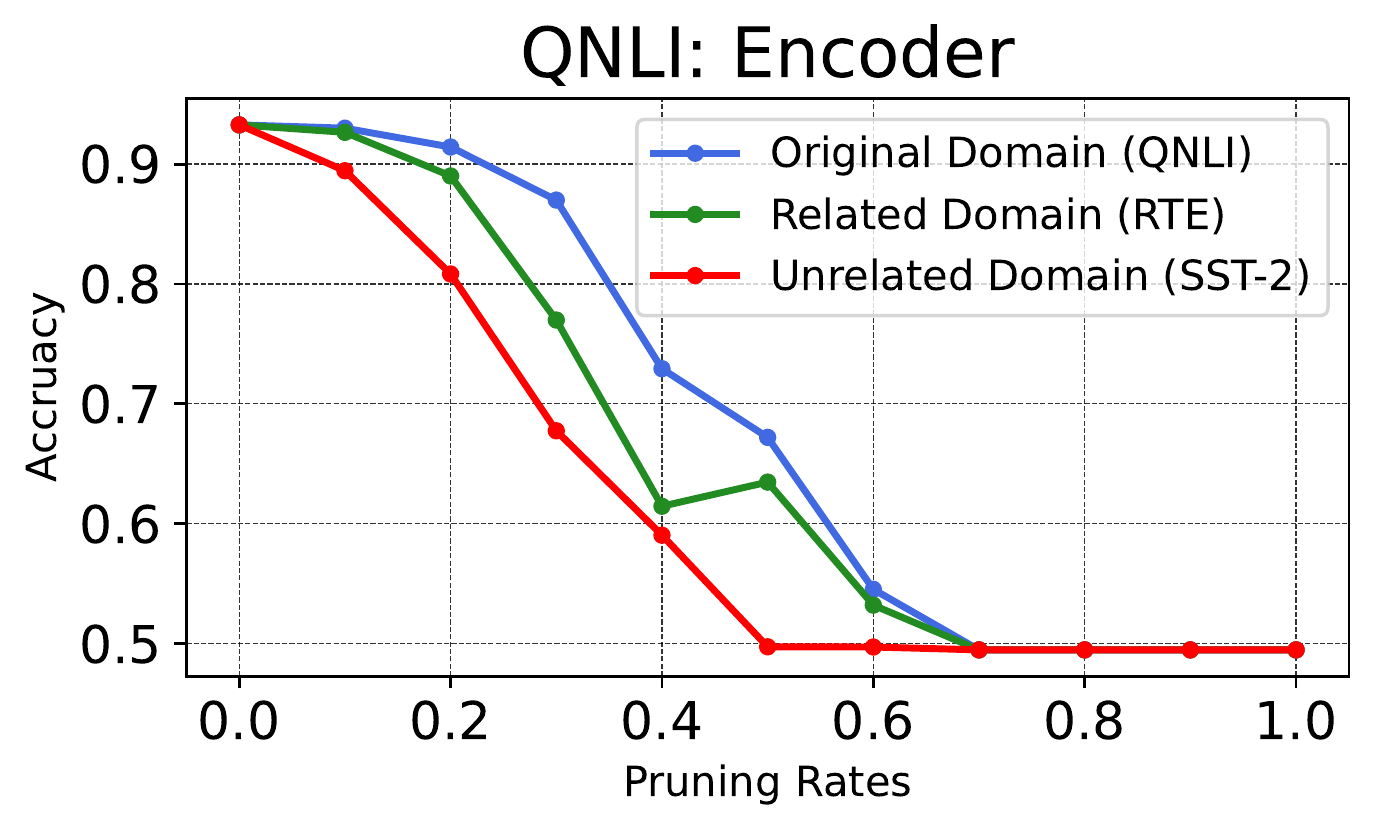}
\end{subfigure}

\begin{subfigure}[t]{0.5\textwidth}
    \centering
    \includegraphics[width=\x cm]{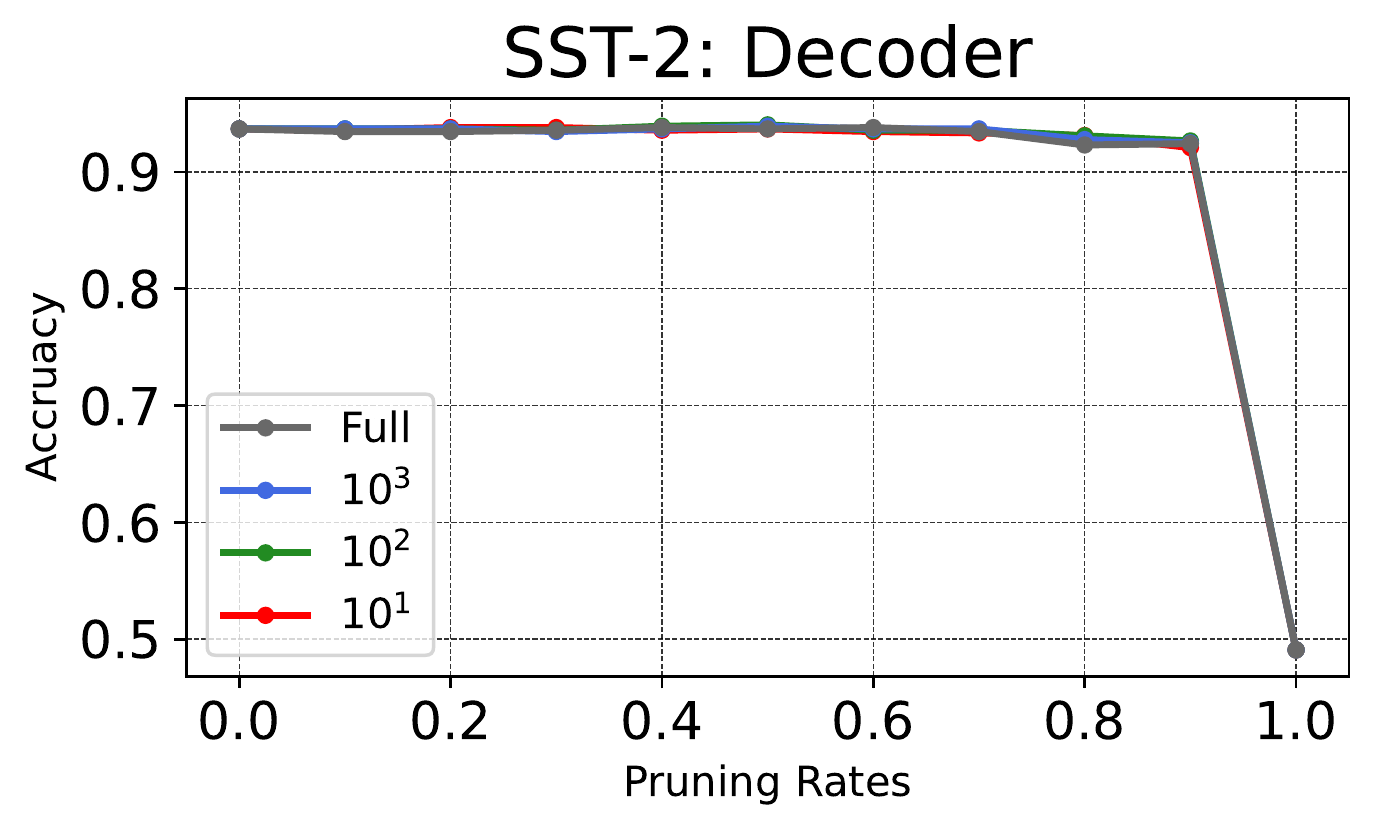}
    \includegraphics[width=\x cm]{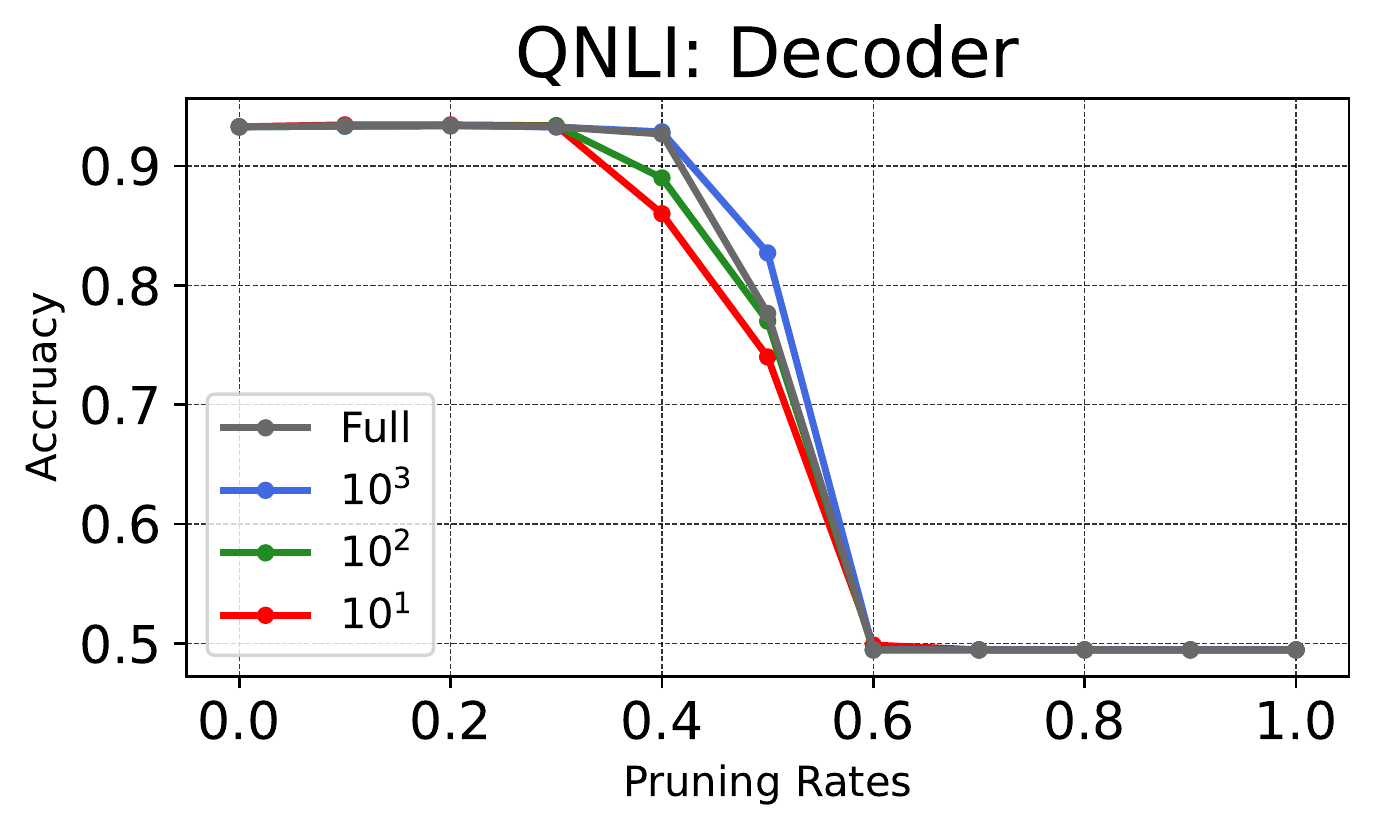}
    \caption{Low Resource Setting}
\end{subfigure}
\begin{subfigure}[t]{0.245\textwidth}
    \centering
    \includegraphics[width=\x cm]{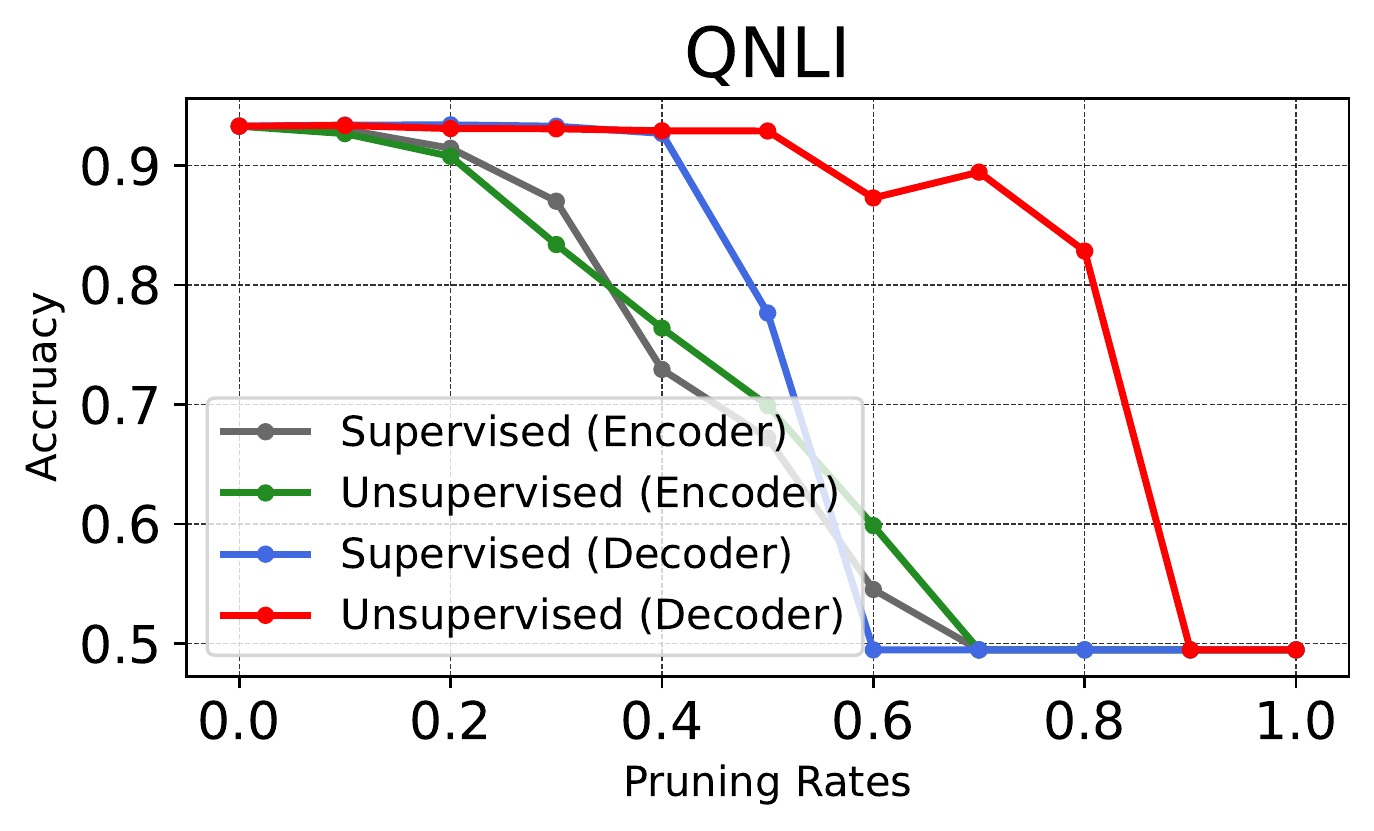}
    \caption{Unsupervised Setting}
\end{subfigure}
\begin{subfigure}[t]{0.245\textwidth}
    \centering
    \includegraphics[width=\x cm]{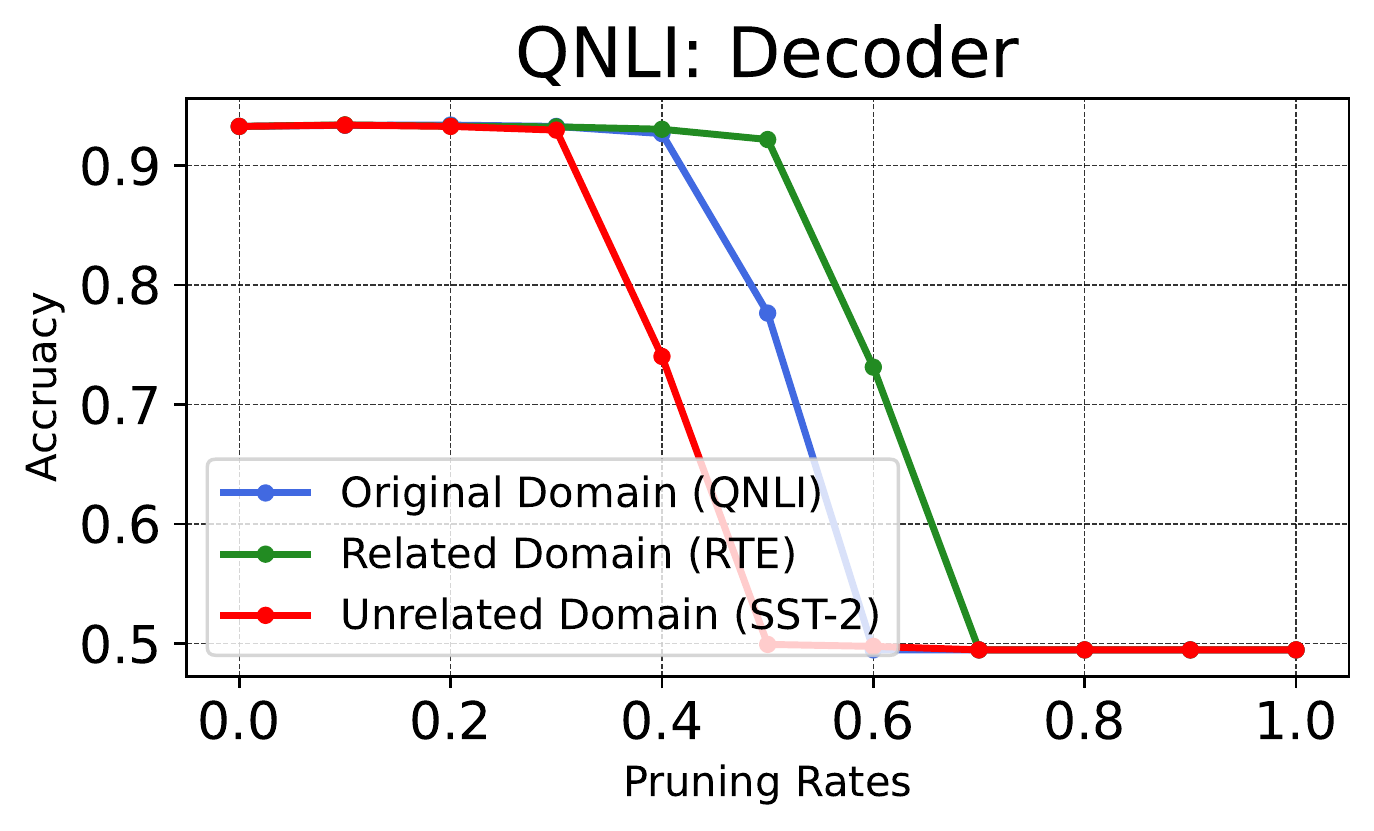}
    \caption{Unseen Domain Setting}
\end{subfigure}

\vspace{-0.2cm}
\caption{Experimental results extending our pruning method to challenging settings: (a) Low-resource setting experiment results. (b) Unsupervised setting experiment results. (c) Unseen domain setting experiment results. These extensions make our method more practical for use in a real-world setting.}
\label{fig:add_exp}
\vspace{-0.47cm}
\end{figure*}

\paragraph{Module-integrated Pruning}
To maximize the compression efficiency of a language model, we should compress the whole model instead of compressing the encoder or decoder, respectively.
Therefore, we also validate our method by compressing the whole architecture of T5.
Figure~\ref{fig:integrated} shows the experimental results of simultaneously compressing both the encoder and decoder using our method.
These experimental results reveal that compressing the whole architecture of the model, not compressing each encoder or decoder separately, does not degrade the model's performance additionally.

Our method focuses on compressing a multi-task language model without any additional training process in a model-agnostic way.
Therefore, it is difficult to compare our method with previous compression research due to the inconsistent experimental setting since previous studies have treated training-based and model-specific compression methods.
Since our method is model-agnostic, it can be utilized broadly and generally to prune multi-task language models containing only task-specific knowledge after applying other compression methods.

\begin{figure*}[!ht]
\newcommand\x{3.3}
\begin{subfigure}[t]{0.49\textwidth}
    \centering
    \includegraphics[width=\x cm]{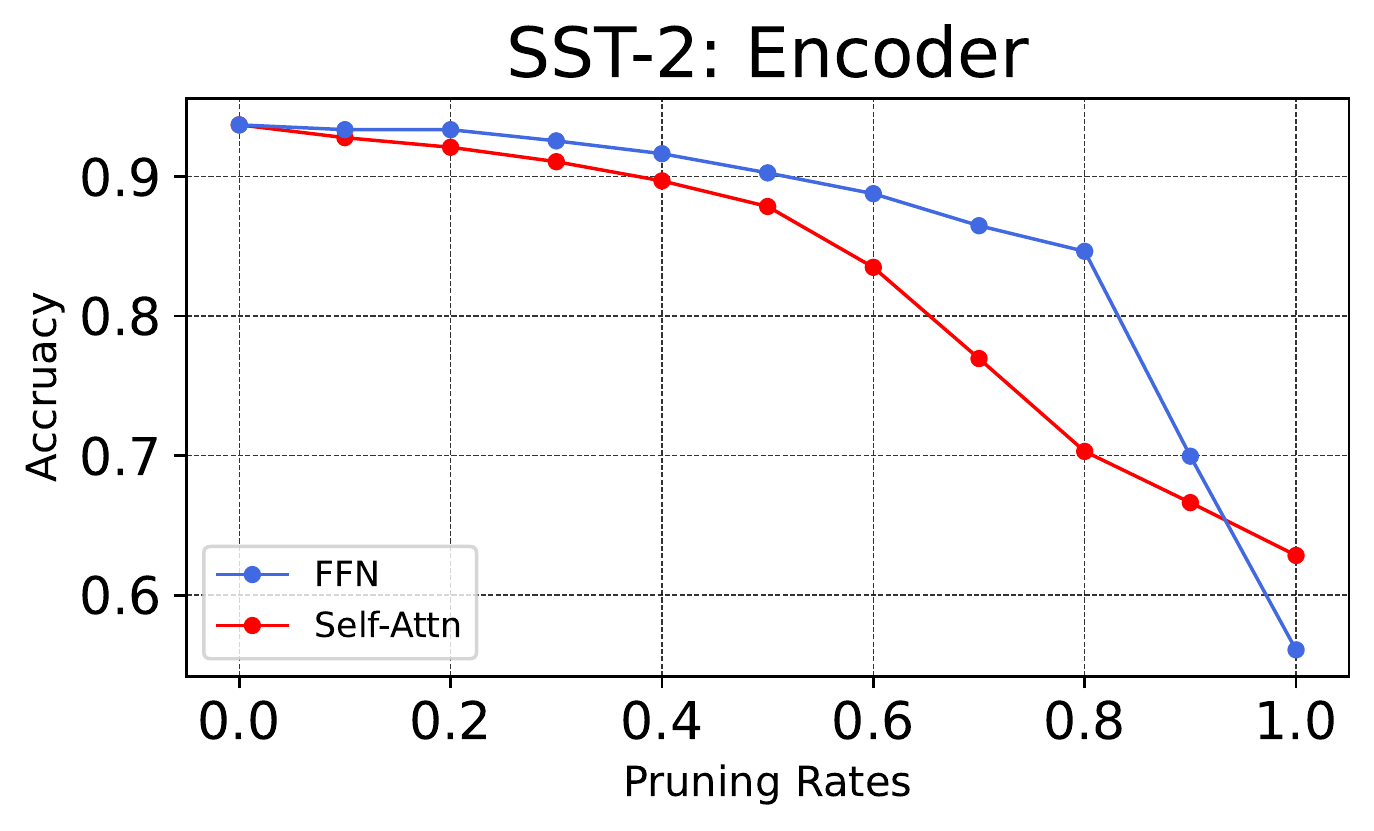}
    \includegraphics[width=\x cm]{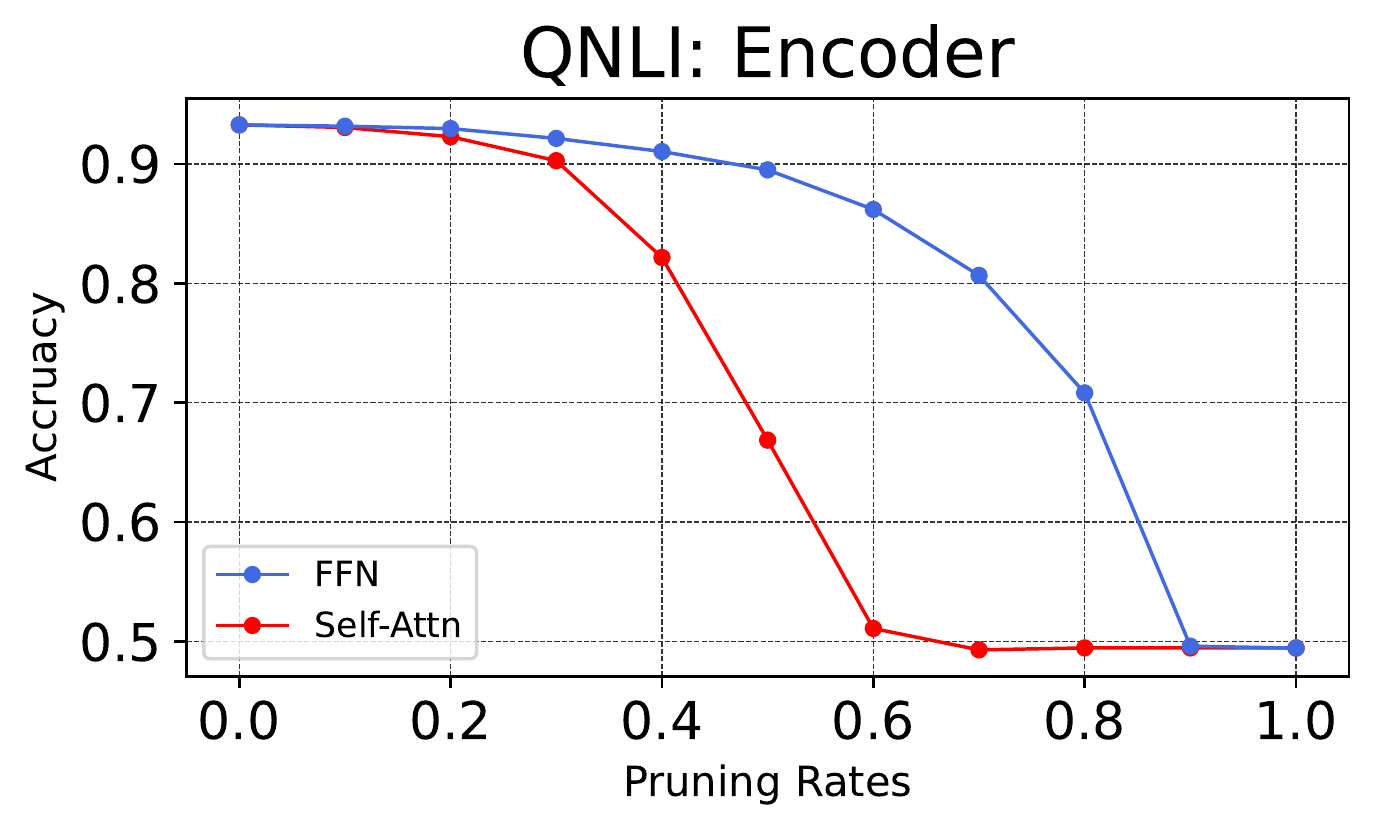}
    \includegraphics[width=\x cm]{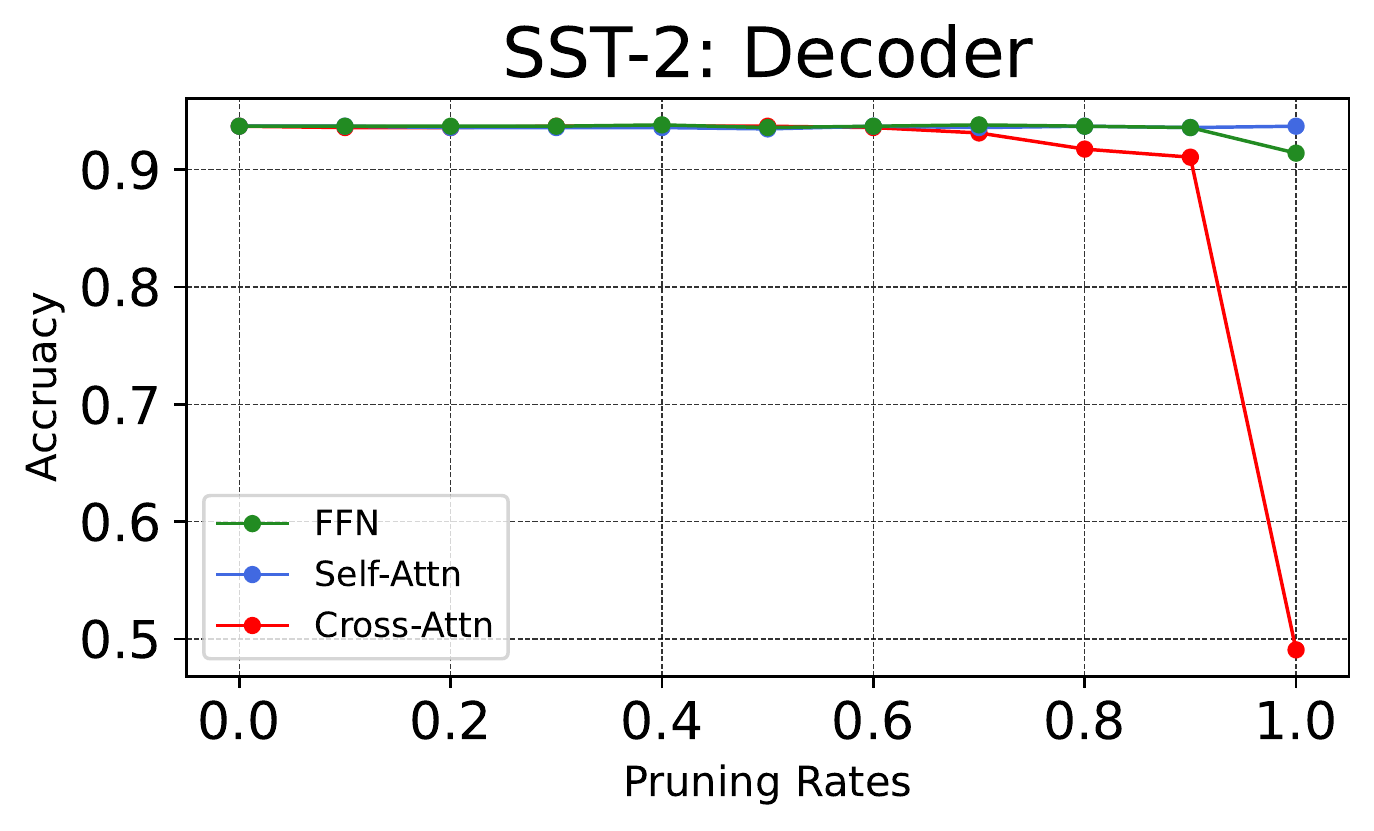}
    \includegraphics[width=\x cm]{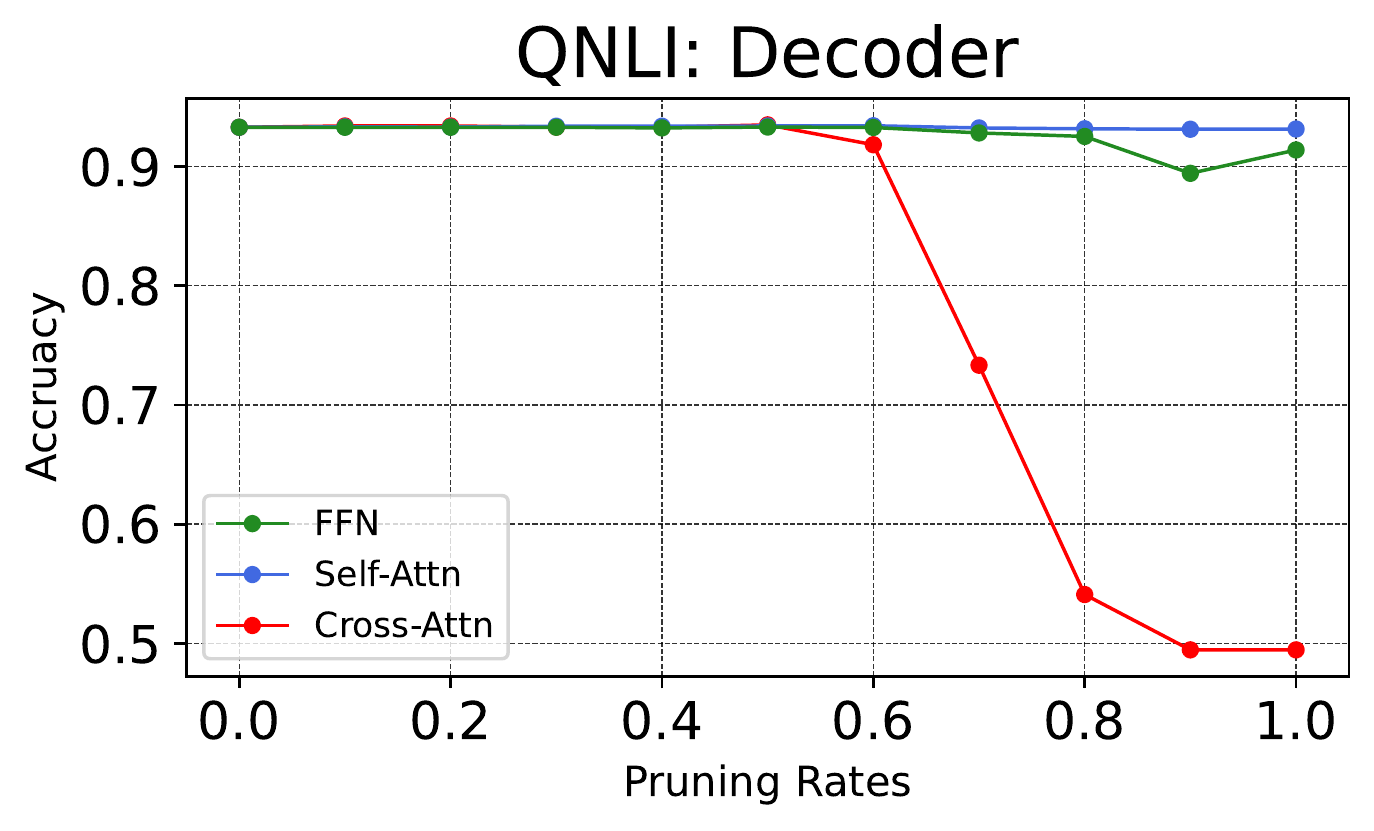}
    \caption{Layer Type-specific Pruning}
\end{subfigure}
\hspace{0.2cm}
\begin{subfigure}[t]{0.49\textwidth}
    \centering
    \includegraphics[width=\x cm]{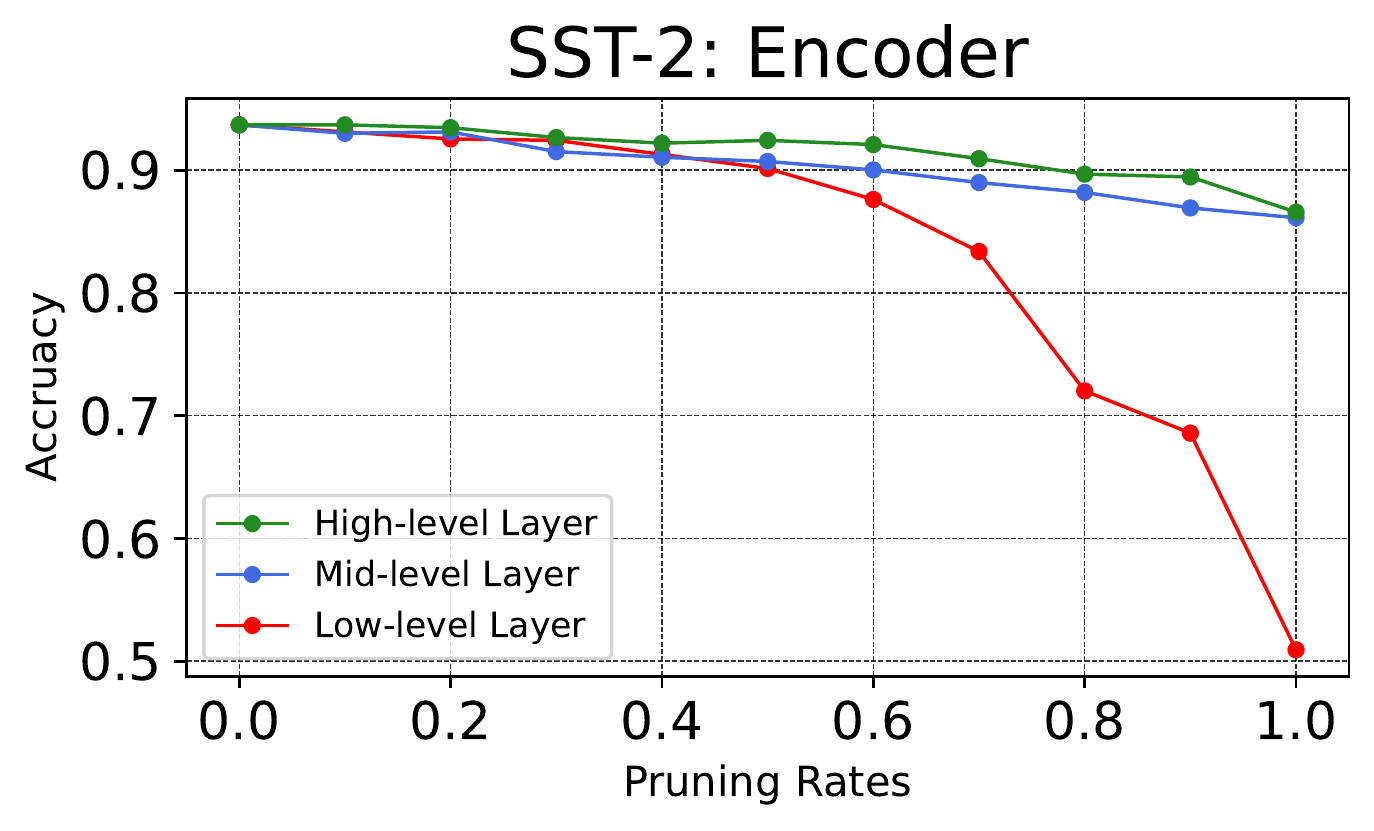}
    \includegraphics[width=\x cm]{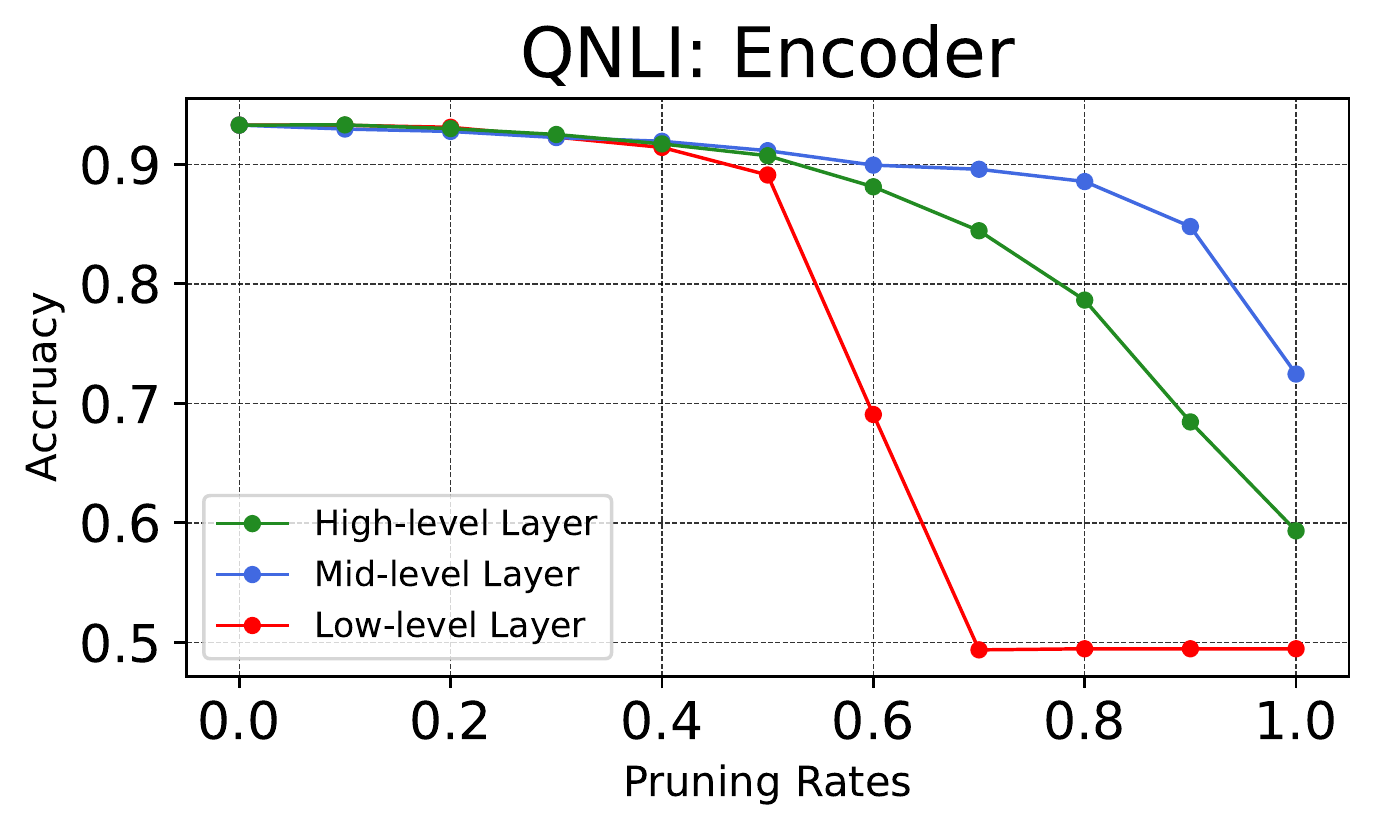}
    \includegraphics[width=\x cm]{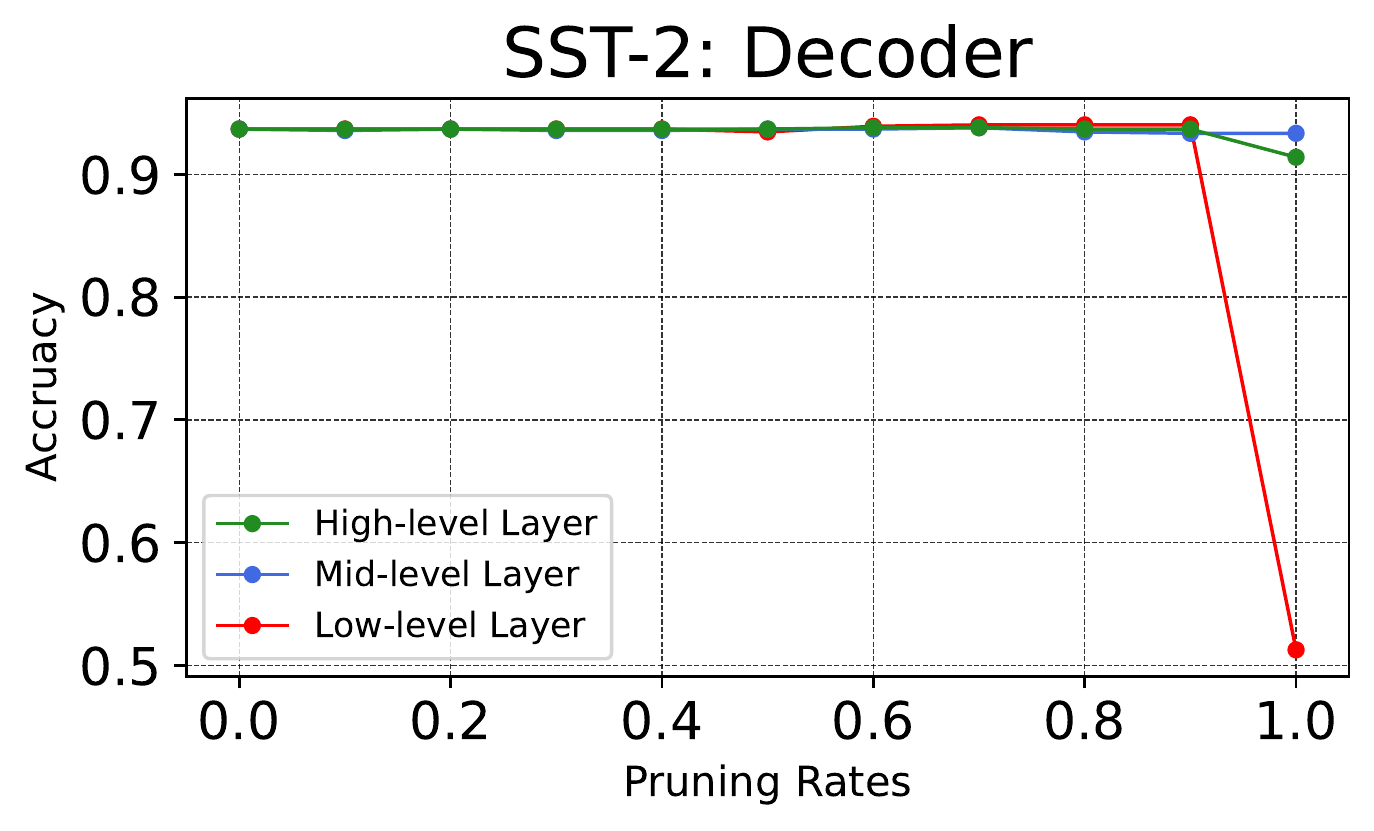}
    \includegraphics[width=\x cm]{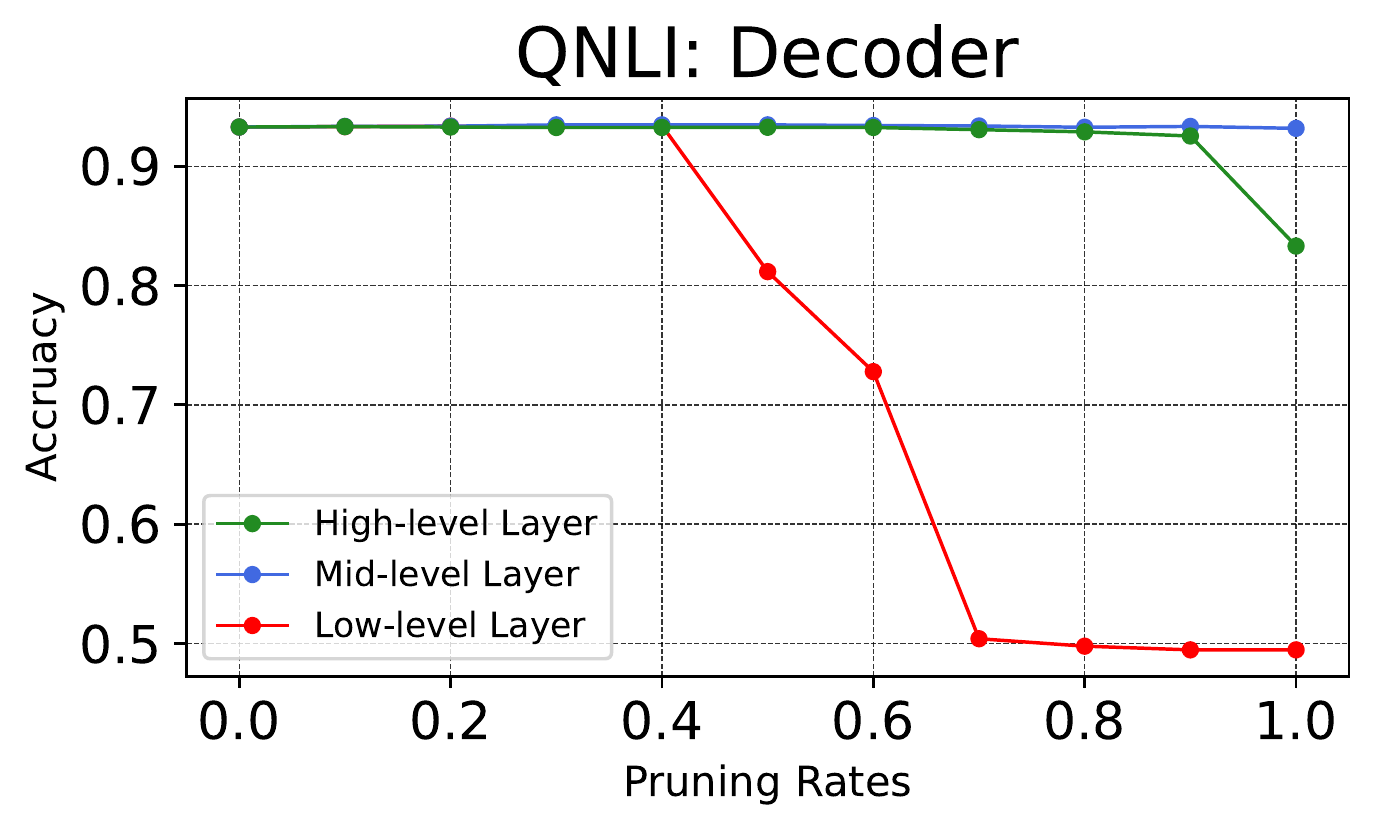}
    \caption{Layer Depth-specific Pruning}
\end{subfigure}

\vspace{-0.2cm}
\caption{Layer types analysis: (a) Layer architecture experiment results. (b) Layer depth experiment results. The higher the degradation, the more essential layers are.}
\label{fig:layer_type_exp}
\vspace{-0.47cm}
\end{figure*}

\subsection{Low-resource Setting}
\label{sec:experiments-low-resource}
In this section, we demonstrate the results for compressing language models based on the attribution computed from only few-shot.
Specifically, we compute neuron importance using only $10^{3}$ and $10^{2}$, and $10^{1}$ samples of SST-2 and QNLI datasets and prune the T5 model with the computed importance, where we balance the number of samples for each class when sampling a subset of the whole dataset.
All results are reported by averaging five trials of random sampling.
Figure~\ref{fig:add_exp}-(a) represents the pruning results in low-resource setting.
For SST-2 dataset, we find that compression using only $10^{1}$ data samples yields comparable performance to the results of using the entire training dataset.
The total number of data samples of SST-2 is 67k, and $10^{1}$ of data samples corresponds to about only $10^{-4}$ of the whole dataset.
For the QNLI dataset, we demonstrate that compression using only $10^{3}$ data samples of the labeled training dataset yields comparable performance to the results of using the entire training dataset.
Furthermore, the performance degradation is also insignificant when using only $10^{1}$ samples of the labeled QNLI training dataset.
The total number of data samples of QNLI is 105k, and $10^{3}$ and $10^{1}$ data samples correspond to about only $10^{-2}$, $10^{-4}$ of the whole dataset, respectively.
These results suggest that most of the task-specific knowledge is derived from computing gradients for only the candidate outputs.
We can effectively reduce the time consumption in this low-resource setting by using a few labeled instances to compute the attribution, and it is the most significant advantage over other training-based compression methods.

\subsection{Unsupervised Setting}
We suggest an additional method to compute attributions using an unlabeled text dataset in section~\ref{sec:method-unsupervised}.
We present the pruning results by computing attributions for an unsupervised setting in Figure~\ref{fig:add_exp}-(b).
Results of encoder compression with the unsupervised setting for both SST-2 and QNLI datasets show competitive scores to that of labeled data.
For the decoder, the performance of SST-2 decreases slightly, but the performance of QNLI rather increases.
The experimental result on SST-2 reveals that the compression in an unsupervised setting shows robust performance maintenance.
In the QNLI result, we observe that computing attributions using information from all output candidates enhances the model's performance.

\subsection{Unseen Domain Setting}
In this section, we validate the effect of our task-specific compression on unseen domains.
We compress the T5 using related and unrelated datasets, and then compare the performance preservation for the original dataset.
Specifically, we compress the T5 using attribution computed with SST-2 and RTE, respectively.
And then, we evaluate the compressed models with the QNLI dataset.
QNLI and RTE are related domains since both are natural language inference datasets, and SST-2 is an unrelated domain built for sentiment analysis.
Figure~\ref{fig:add_exp}-(c) shows the evaluation results of the compressed model for related and unrelated domains.
Experimental results reveal our method's robust performance maintenance for the related domain.
Surprisingly the case of decoder compression shows even better performance maintenance in the related domain than in the original domain.

\subsection{Layer-specific Pruning Analysis}
This section further investigates the pruning effect per layer type.
We select two pruning settings: (1) Layer type-specific and (2) Layer depth-specific.

\paragraph{Layer Type-specific Pruning Analysis} Layer type-specific pruning analysis focuses on understanding how the performance of the model varies depending on the type of compressed layers.
The encoder investigates pruning results for feed-forward neural networks and self-attention networks, and the decoder focuses on feed-forward neural networks, self-attention networks, and cross-attention networks.

\paragraph{Layer Depth-specific Pruning Analysis} Layer depth-specific pruning analysis investigates how the performance of the model changes depending on the depth of the compressed layers.
We select SST-2 for experiments and separate each encoder and decoder into three parts: (1) Low-level layer, (2) Mid-level layer, and (3) High-level layer.
Since \emph{T5-base} consists of 12 layers for each encoder and decoder, each depth consists of 4 layers.

Layer-specific pruning results are shown in Figure~\ref{fig:layer_type_exp}.
For the encoder, self-attention networks are more critical for preserving the performance than feed-forward neural networks.
For the decoder, cross-attention networks are more important than feed-forward neural networks and self-attention networks.
For each layer-depth, we can conclude that the low-level features are more crucial to preserving the model's performance.
Especially, experimental results reveal that the model's performance is preserved even if the pruning rate of a specific layer is 1.0.
These results demonstrate that there is redundant information processing between layers for performing a specific task.
Note that although the pruning rate is 1.0 for a layer, the representation propagated through the pruned layer does not lose every knowledge completely.
It is because transformer variants have residual connections to preserve the knowledge of previous layers.

\section{Conclusion}
This paper proposes a novel training-free attribution-based task-specific knowledge extraction method for multi-task language models.
Specifically, we use attribution to determine which neurons are important to derive a specific output for each task.
Then, we prune task-specifically unimportant neurons to extract only task-specific knowledge from the entire model.
We further propose a method for computing attributions in low-resource and unsupervised settings.
We demonstrate that our method outperforms the other pruning methods on the widely used text datasets.
In addition, we examine that our task-specific language model pruning method shows outstanding performance in the unseen domain, especially when the unseen domain is related to the dataset used to configure the compressed version.
Our compression method does not update the pre-trained parameters of the language models, which enables efficient on-demand compression and inference.
Also, our proposed method is valuable because it can be universally applied to any neural network-based model architecture.

\section*{Limitations}
To the best our knowledge, this is the first work to compress a multi-task language model without extra training on the target task.
Due to insufficient prior work on these training-free compression methods, we couldn't include a thorough comparison with other baseline algorithms.
Also, our work focused on analyzing the results of six widely-used natural language understanding datasets among GLUE benchmark.
We believe that extra experiments on various challenging natural language understanding tasks will show our work's generalization performance. We have conducted experiments on various settings; varying layer types, layer depth, low resource, unsupervised, and unseen domain.
However, there are still extra room for improving this work, such as exploring and applying layer-specific pruning rates, which we leave for future work. 

\section*{Acknowledgements}
We thank anonymous reviewers for their constructive and insightful comments.
K. Jung is with ASRI, Seoul National University, Korea. 
This work was supported by AIRS Company in Hyundai Motor Company \& Kia Motors Corporation through HMC/KIA-SNU AI Consortium Fund.
This work was partly supported by Institute of Information \& communications Technology Planning \&
Evaluation (IITP) grant funded by the Korea government(MSIT) [NO.2022-0-00184, Development and Study of AI Technologies to Inexpensively Conform to Evolving Policy on Ethics \& NO.2021-0-02068, Artificial Intelligence Innovation Hub (Artificial Intelligence Institute, Seoul National University) \& NO.2021-0-01343, Artificial Intelligence Graduate School Program (Seoul National University)]

\bibliography{anthology,custom}
\bibliographystyle{acl_natbib}

\appendix
\section{Algorithms}\label{sec:appendix1}
Our pruning method consists of two stages: (1) Derivation of task-specific neuron indices per layer for a specific task (2) Real-time task-specific inference with previously pruned layers.

\algrenewcommand\algorithmicindent{1.5em}%
\algrenewcommand\algorithmicrequire{\textbf{Input:}}
\algrenewcommand\algorithmicensure{\textbf{Output:}}
\begin{algorithm}
 \caption{Deriving task-specific neuron indices per layer for a task $t$}\label{alg:cap}
\begin{algorithmic}[1]
\small 
\Require task-specific dataset $\mathcal{D}^{t}$; model $\mathcal{P}$; pruning rate $p$
\Ensure list $\mathcal{M}^{t}$ with task-specific neuron indices per layer \\
Initialize all $\mathcal{M}_{l}^{t}$ as an empty set and all $A^{(\mathcal{D}^{t})}_{i}$ to zero
\State $\mathcal{B} \gets $ split $\mathcal{D}^{t}$ into mini-batches of size $\beta$
\For{each batch $b \in \mathcal{B}$}
    \For{each layer $l \in \mathcal{P}$}
        \For{ $i=1$ to $k^{l}$}
            \vspace{-0.1cm}
            \State compute neuron importance $A^{(b)}_{i}(h^{l})$
            \State $A^{(\mathcal{D}^{t})}_{i}(h^{l}) \gets A^{(\mathcal{D}^{t})}_{i}(h^{l}) + A^{(b)}_{i}(h^{l})$
        \EndFor
    \EndFor
\EndFor

\For {each layer $l \in \mathcal{P}$}
    \For {$i=1 $  to $k^{l}$}
        \If{$argsort_i(A^{(\mathcal{D}^{t})}(h^{l})) < \lfloor k^{l}\times p \rfloor $ }
            \vspace{0.03cm}
            \State $\mathcal{M}_{l}^{t} \gets \mathcal{M}_{l}^{t} \cup \{i\}$
        \EndIf
    \EndFor
\EndFor
\Return $\mathcal{M}^{t}$
\end{algorithmic}
\label{algorithm:derive-mask}
\vspace{-0.02cm}
\end{algorithm}

\vspace{-0.3cm}
In the first stage, we sort neuron indices in descending order by computed attribution scores, leaving high-importance neurons by $(1-p)$ ratio.

\algrenewcommand\algorithmicindent{0.9em}%
\algrenewcommand\algorithmicrequire{\textbf{Input:}}
\algrenewcommand\algorithmicensure{\textbf{Output:}}
\begin{algorithm}
\caption{Real-time task-specific inference with pruned layers}\label{alg:cap}
\begin{algorithmic}[1]
\small 
\Require task $t$; text inputs $x$; indices container $\mathcal{M}$; model $\mathcal{P}$
\Ensure text outputs $y$
\State For task $t$, load corresponding $\mathcal{M}^{t}$
\For {each layer $l \in \mathcal{P}$} 
  \State $W^{l} \gets (W_{ij}^{l})_{\substack{i \in \mathcal{M}^{t}_{l'}\\ j\in \mathcal{M}^{t}_{l}}}$ \Comment{\scriptsize{match rows with a previous layer $l'$}}
  \algrenewcommand\algorithmicindent{1.5em}%
  \small 
  \If{ bias $b^{l}$ exists in layer $l$}
    \State $b^{l} \gets (b_{i}^{l})_{i \in \mathcal{M}^{t}_{l}}$
  \EndIf
\EndFor
\vspace{-0.1cm}
\State compute outputs $y$ with $x$ using the pruned model $\tilde{\mathcal{P}}$

\noindent\Return $y$
\end{algorithmic}
\label{algorithm:inference}
\end{algorithm}

\vspace{-0.3cm}
In the second stage, we prune task-specifically unimportant neurons when given a user request for a specific task. 
We task-specifically compress a model in real-time and conduct an inference with the pruned model.

\section{Statistic of Pruning Results} \label{sec:appendix2}
We compute the pruning results of the baselines of T5-RP and T5-RAP through five random trials.
The standard deviations of the accuracy for the two baselines are shown in Table~\ref{table:std}.

\begin{table}[ht]
\small
\centering
\resizebox{\linewidth}{!}{\begin{tabular}{cccccccc}
\hline
\textbf{} & \textbf{} & \textbf{SST-2} & \textbf{MRPC} & \textbf{QNLI} & \textbf{RTE} & \textbf{CB} & \textbf{BoolQ} \\ 
\hline
\multirow{2}{*}{\textbf{T5-RP}} & \textbf{Encoder} & 0.0202 & 0.0046 & 0.0143 & 0.0426 & 0.0168 & 0.0020 \\ 
 & \textbf{Decoder} & 0.0241 & 0.0108 & 0.0003 & 0.0580 & 0.0010 & 0.0060 \\
\hline
\multirow{2}{*}{\textbf{T5-RAP}} & \textbf{Encoder} & 0.0082 & 0.0080 & 0.0119 & 0.0165 & 0.0099 & 0.0061 \\ 
 & \textbf{Decoder} & 0.0159 & 0.0089 & 0.0029 & 0.0123 & 0.0027 & 0.0063 \\
\hline
\end{tabular}}
\vspace{-0.3cm}
\caption{
Standard deviations of Pruning results.
}
\vspace{-0.2cm}
\label{table:std}
\end{table}

We calculate the standard deviations by averaging the values derived by all pruning rates.
These results reveal that the variances of T5-RP and T5-RAP are not significant.

\end{document}